\documentclass[sigconf,screen,nonacm]{acmart}
\AtBeginDocument{%
  }

\setcopyright{none}
\hypersetup{
  pdfauthor={Ruoran Xu, Wending Gao, Qiufeng Wang},
  pdftitle={FormalAnalyticGeo: A Neural-Symbolic Based Framework for Multimodal Analytic Geometry Problem Generation}
}

\usepackage[table]{xcolor}
\usepackage{float}
\usepackage{tcolorbox}
\tcbuselibrary{listings,breakable,skins}

\definecolor{promptheader}{RGB}{0,70,140}       
\definecolor{promptwarn}{RGB}{180,30,30}         
\definecolor{prompttool}{RGB}{0,120,120}         
\definecolor{promptcdl}{RGB}{100,60,150}         

\lstdefinestyle{promptstyle}{
  basicstyle=\ttfamily\scriptsize,
  breaklines=true,
  columns=fullflexible,
  keepspaces=true,
  aboveskip=0pt,
  belowskip=0pt,
  alsoletter={_},
  morecomment=[l][\color{promptheader}\bfseries]{\#\#},
  morekeywords={[2]CRITICAL,IMPORTANT,NEVER,MANDATORY,MUST,ALWAYS,INVALID,PROHIBITIONS},
  keywordstyle={[2]\color{promptwarn}\bfseries},
  morekeywords={[3]seed_pool,enumerate_properties,sympy_check,rule_check,submit_problem,submit_cdl,pixel_to_coord,coord_to_pixel,find_nearest_curve_point,compute_distance,compute_slope,compute_angle,compute_area,scan_region,profile_line,intersect_line_curve,intersect_curves,verify_point_on_curve,check_answer_range,validate_cdl_syntax,check_cdl_completeness,verify_solvability,solve_text_only},
  keywordstyle={[3]\color{prompttool}},
  morekeywords={[4]PointOnCurve,Focus,Vertex,Center,Directrix,Asymptote,MidPoint,Intersection,TangentOfPoint,TangentOnPoint,TangentPoint,Projection,RightPart,LeftPart,OverlappingLine,LineSegmentOf,TriangleOf,VectorOf,DotProduct,Equation,Coordinate,IsPerpendicular,IsParallel,IsTangent,IsChordOf,IsDiameter,Area,Distance,Slope,Eccentricity,FocalLength,Inclination,Length,SemiMajorAxis},
  keywordstyle={[4]\color{promptcdl}},
  moredelim=[is][\bfseries]{**}{**},
}

\definecolor{pykw}{RGB}{0,0,180}         
\definecolor{pystr}{RGB}{163,21,21}      
\definecolor{pycmt}{RGB}{0,128,0}        
\definecolor{pytype}{RGB}{0,120,120}     
\definecolor{pyfunc}{RGB}{120,50,0}      
\lstdefinestyle{pythonstyle}{
  language=Python,
  basicstyle=\ttfamily\scriptsize,
  breaklines=true,
  columns=fullflexible,
  keepspaces=true,
  aboveskip=0pt,
  belowskip=0pt,
  keywordstyle=\color{pykw}\bfseries,
  stringstyle=\color{pystr},
  commentstyle=\color{pycmt}\itshape,
  emphstyle=\color{pyfunc}\bfseries,
  emph={pixel_to_coord,coord_to_pixel,find_nearest_curve_point,compute_distance,compute_slope,compute_angle,compute_area,scan_region,profile_line,intersect_line_curve,intersect_curves,verify_point_on_curve},
  morekeywords={dict,int,float,str,list,bool,Optional},
  showstringspaces=false,
}
\newtcblisting{pythonbox}[1]{
  title={#1},
  colback=white,
  colframe=black,
  coltitle=white,
  colbacktitle=black,
  fonttitle=\bfseries\small,
  listing only,
  listing options={style=pythonstyle},
  breakable,
  enhanced jigsaw,
  left=2mm, right=2mm, top=1mm, bottom=1mm
}

\definecolor{cdlgreen}{RGB}{0,120,60}
\lstdefinestyle{cdlstyle}{
  basicstyle=\ttfamily\small\color{cdlgreen},
  breaklines=true,
  columns=fullflexible,
  keepspaces=true,
  aboveskip=4pt,
  belowskip=4pt,
}
\lstnewenvironment{cdlcode}{\lstset{style=cdlstyle}}{}

\newtcblisting{promptbox}[1]{
  title={#1},
  colback=white,
  colframe=black,
  coltitle=white,
  colbacktitle=black,
  fonttitle=\bfseries\small,
  listing only,
  listing options={style=promptstyle},
  breakable,
  enhanced jigsaw,
  left=2mm, right=2mm, top=1mm, bottom=1mm
}

\newcommand{\blfootnote}[1]{%
  \begingroup
  \renewcommand{\thefootnote}{}%
  \begin{NoHyper}%
  \footnotetext{#1}%
  \end{NoHyper}%
  \addtocounter{footnote}{-1}%
  \endgroup
}

\begin{document}

\title{FormalAnalyticGeo: A Neural-Symbolic Based Framework for Multimodal Analytic Geometry Problem Generation}

\author{\texorpdfstring{%
  Ruoran Xu\textsuperscript{*}\quad
  Wending Gao\textsuperscript{*}\quad
  Xiaoqiang Kang\quad
  Qiufeng Wang\textsuperscript{\textdagger}%
}{Ruoran Xu, Wending Gao, Qiufeng Wang}}
\renewcommand{\shortauthors}{Xu, Gao, and Wang}

\begin{abstract}
Math reasoning has achieved significant progress with the rapid advancement of Multimodal Large Language Models (MLLMs), however analytic geometry remains largely underexplored, primarily due to the scarcity of annotated 
samples. 

Existing diagram generation approaches struggle with analytic geometry: template methods cannot handle constraint-driven layouts, and generative models lack the geometric precision to render annotated conic curves correctly.

We present FormalAnalyticGeo, a scalable framework for fully automatic generation of multimodal analytic geometry problems. Leveraging the rigor of formal languages, we design the framework around CDL (Condition Description Language), a formal intermediate representation that bridges free-form problem text with precise diagram rendering via a Signed Distance Field (SDF) engine. The framework employs four specialized LLM components in sequence: a Generator that produces diverse analytic geometry problems, a Formalizer that converts each problem into CDL for SDF-based rendering, a Measurer that extracts ground-truth answers through vision-based measurement on the rendered diagrams, and a Quality Verifier that checks outputs at three stages. Structured feedback from the Quality Verifier drives automatic retry, forming a closed loop that eliminates any need for human annotation.

Applying FormalAnalyticGeo at scale yields AnalyticGeo7K, a dataset of over 7K verified multimodal problems, each with aligned text, diagram, formal annotation, and ground truth.Experiments show that the generated problems achieve a median ground-truth relative error of 0.70\%, with 82.3\% of answers falling within 5\% of the exact symbolic solution. Our framework and dataset will be publicly released.

\end{abstract}

\maketitle
\blfootnote{%
Xi'an Jiaotong-Liverpool University\\
\textsuperscript{*} Equal contribution.\quad
\textsuperscript{\textdagger} Corresponding author: qiufeng.wang@xjtlu.edu.cn
}



\section{Introduction}



Multimodal Large Language Models (MLLMs) have exhibited impressive performance in mathematical reasoning. However, most existing research on multimodal mathematical reasoning primarily focuses on elementary plane geometry and solid geometry, leaving analytic geometry as a largely underexplored domain, while the capacity to jointly comprehend visual diagrams and complex algebraic structures continues to pose a considerable challenge, it demands an integrated understanding of equations, coordinate systems, geometric figures including conic sections, and the intricate spatial relationships between them.

Despite its importance in both education and standardized examinations,
the multimodal understanding of analytic geometry by MLLMs remains largely underexplored,
in part because existing geometry benchmarks address plane geometry exclusively
and the only large-scale conic-section resource (Conic10K~\cite{wu-etal-2023-conic10k}) is purely text-based.
Automated data generation is therefore essential,
yet building a reliable generation framework for analytic geometry
poses methodological challenges that existing approaches have not addressed.
\begin{figure}[H]
    \centering
    \includegraphics[width=1\linewidth]{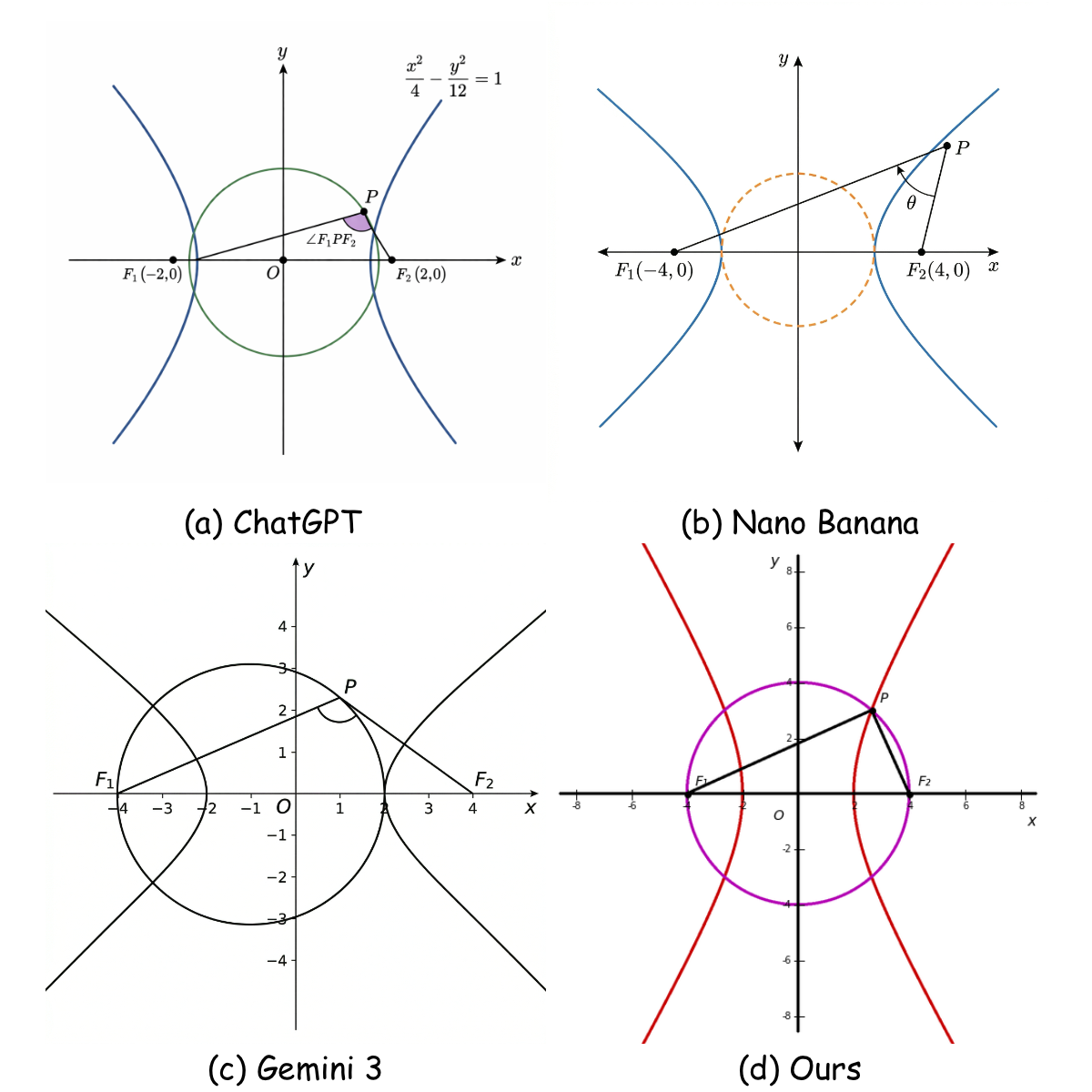}
    \caption{Diagram generation for the same analytic geometry problem. (a) ChatGPT: the circle does not use the focal distance as its diameter, and $P$ lies on the circle but not on the hyperbola. (b) Nano Banana: geometrically meaningless output. (c) Gemini 3 Pro: distorted curves and inaccurate point placement. (d) Ours: geometrically exact rendering with known coordinate-to-pixel mapping, enabling direct visual measurement of ground-truth answers.}
    \label{fig:placeholder}
\end{figure}
We identify three such challenges.
\textbf{(i)~Representation gap.}
Generating multimodal problems requires a formal representation
that captures both algebraic relationships and geometric constructions
needed to render a diagram.
Existing geometry formalisms (Inter-GPS~\cite{lu2021inter}, FormalGeo~\cite{zhang2024formalgeoextensibleformalizedframework})
target plane geometry and cannot express coordinate systems or conic curves;
Conic10K's~\cite{wu-etal-2023-conic10k} Assertional Logic encodes algebraic relations
but lacks the construction primitives required for rendering.
\textbf{(ii)~Rendering gap.}
Analytic geometry diagrams involve algebraically defined curves
embedded in a coordinate system.
Existing tools such as GeoGPT4V~\cite{cai-etal-2024-geogpt4v} are restricted to Euclidean constructions;
MAVIS~\cite{zhang2024mavismathematicalvisualinstruction} spans analytic geometry but lacks formal text--diagram verification.
Traditional plotting libraries require all positions to be pre-computed,
offering no mechanism for constraint-driven layout.
\textbf{(iii)~Verification gap.}
Existing generation approaches are single-pass,
accepting outputs without cross-stage verification;
errors at any stage propagate undetected into the final dataset.

We propose FormalAnalyticGeo,
a scalable framework that addresses all three challenges through formal language.
To close the \emph{representation gap},
we design CDL (Condition Description Language),
a formal intermediate representation that captures both algebraic relations
and geometric constructions, enabling automatic consistency checking.
To close the \emph{rendering gap},
we develop an SDF-based rendering engine
that compiles CDL programs into differentiable Signed Distance Fields,
unifying constraint solving and diagram rendering
so that under-determined positions are resolved via gradient descent.
To close the \emph{verification gap},
a Quality Verifier checks each stage's output at three gate positions
and returns structured feedback for automatic retry,
forming a closed loop that eliminates any need for human annotation.
The framework comprises four stages:
a Generator produces diverse problems,
a Formalizer translates each into CDL,
the SDF engine renders a diagram,
and a Measurer extracts ground-truth answers
through vision-based measurement on the rendered diagram.
Applying FormalAnalyticGeo at scale yields over 7K verified problems,
each comprising a natural language question, a standardized diagram,
a CDL annotation, and a visually extracted ground-truth answer,
with a median relative error of 0.70\%.

Our main contributions are as follows:
\begin{itemize}
    \item \textbf{A closed-loop generation framework}: FormalAnalyticGeo decomposes multimodal analytic geometry data generation
    into specialized stages connected by a formal language,
    with feedback-driven quality gates eliminating the need for human annotation.

    \item \textbf{A formal language for analytic geometry}: CDL bridges natural language problem descriptions and geometric rendering,
    enabling automatic syntactic validation and completeness checking.

    \item \textbf{An SDF-based rendering engine}: SDF Engine compiles CDL into differentiable Signed Distance Fields,
    unifying constraint solving and diagram rendering for exact analytic curves.

    \item \textbf{A large-scale multimodal dataset}: AnalyticGeo7k as a product of the framework, it covers over 7K verified problems with ablation studies validating each component.

    \item \textbf{Comprehensive Empirical Analysis}: We conduct comprehensive experiments on 8 state-of-the-art models and providing actionable insights for advancing multimodal spatial and mathematical reasoning.
    
\end{itemize}

\section{Related Work}
\label{sec:2}
We review related work in three areas: multimodal mathematical reasoning benchmarks, geometry data generation, and multi-agent systems and tool use.

\subsection{Multimodal Math Reasoning Benchmarks}
\label{sec:2.1}
Recent benchmarks have driven rapid progress in multimodal mathematical reasoning.
MathVista~\cite{lu2024mathvista} consolidates 28 existing datasets into a 6,141-problem suite
and shows GPT-4V~\cite{openai2024gpt4technicalreport} achieves only 49.9\% versus 60.3\% for humans.
MathVerse~\cite{zhang2024mathverse} further reveals that some MLLMs perform \emph{better}
when diagrams are withheld, indicating that models routinely ignore visual content.
MATH-Vision~\cite{wang2024measuring} extends this to 3,040 competition-level problems across
16 subjects including analytic geometry, with GPT-4V~\cite{openai2024gpt4technicalreport} scoring only 23.98\%.
DynaMath~\cite{zou2024dynamic} constructs 5,010 variants from 501 seed problems
to evaluate VLM robustness under systematic visual perturbations.

For plane geometry, GeoQA~\cite{chen-etal-2021-geoqa}, UniGeo~\cite{chen2022unigeounifyinggeometrylogical}, Geometry3K~\cite{lu2021inter}, Hilbert-Geo~\cite{xu2026hilbertgeo}, and PGPS9K~\cite{Zhang2023PGPS}
together supply over 30,000 problems focused on Euclidean constructions
(triangles, circles, and quadrilaterals), accompanied by symbolic annotations and program solutions.

Analytic geometry remains largely uncovered.
GeoEval~\cite{zhang-etal-2024-geoeval} explicitly includes analytic geometry
but contributes fewer than 1\% of its corpus.\footnote{GeoEval contains 28 analytic geometry problems out of 5,050 total.}
Conic10K~\cite{wu-etal-2023-conic10k} is the only large-scale conic-section dataset (10,861 problems),
with formal AL annotations and multi-step reasoning chains,
but it is purely text-based: it provides neither rendered diagrams
nor a multimodal evaluation protocol.
FormalAnalyticGeo directly addresses this gap by providing an annotation-free generation framework
that pairs each problem with a standardized, SDF-rendered diagram.

\subsection{Geometry Data Generation}
\label{sec:2.2}
Automated geometry data generation has taken several forms.
AlphaGeometry~\cite{AlphaGeometryTrinh2024} synthesizes 100 million proof traces
by combining a neural language model with a symbolic deduction engine (DDAR),
achieving gold-medalist performance on IMO plane-geometry problems.
R-CoT~\cite{linger-etal-2025-theorem} proposes a two-stage forward-then-reverse pipeline
that generates diagram descriptions before deriving problems from them.
These approaches are confined to Euclidean proof tasks
and produce no multimodal pairs suitable for algebraic or metric reasoning.

More recent work targets multimodal geometry data directly.
GeoGPT4V~\cite{cai-etal-2024-geogpt4v} uses GPT-4~\cite{openai2024gpt4technicalreport} and GPT-4V~\cite{openai2024gpt4technicalreport} with Wolfram Alpha
to generate 4.9K plane-geometry problems with diagram images.
G-LLaVA~\cite{gao2025gllava} converts formal logic representations to Geo170K~\cite{gao2025gllava} multimodal pairs,
while TrustGeoGen~\cite{fu2026trustgeogenformalverifieddataengine} introduces formal-verification guarantees
for a Euclidean geometry data engine.
NeSyGeo~\cite{wu2025nesygeoneurosymbolicframeworkmultimodal} proposes a neuro-symbolic framework that defines an entity-attribute-relation DSL
and combines symbolic construction with LLM-based question generation
to synthesize 100K plane-geometry caption and reasoning samples.
AutoGeo~\cite{10960701} and the concurrent Socratic-Geo~\cite{jiao2026socraticgeosyntheticdatageneration}
further demonstrate multi-agent pipelines for plane-geometry diagram synthesis.
Despite their advances, all of these systems target Euclidean or plane geometry exclusively.

Geometry formal languages and formalization methods underpin rendering and reasoning pipelines.
Inter-GPS~\cite{lu2021inter} defines 91 geometric predicates for plane-figure logic forms;
FormalGeo~\cite{zhang2024formalgeoextensibleformalizedframework} extends this to 88 predicates and 196 theorems.
AutoGPS~\cite{ping2026autogps} introduces a Multimodal Problem Formalizer that extracts formal representations
from geometry diagrams and text, feeding a Deductive Symbolic Reasoner
for interpretable, step-by-step problem solving.
Conic10K's~\cite{wu-etal-2023-conic10k} AL formalizes conic-section problems with 90 operators,
but AL encodes only algebraic relations and lacks the geometric construction information
required to render diagrams (e.g., explicit point positions, intersections, tangent lines).
Our CDL extends AL with construction primitives that drive SDF-based rendering.
To the best of our knowledge, no prior work has addressed
analytic geometry data generation for multimodal benchmarks.

\subsection{LLM Agents and Tool Use}
\label{sec:2.3}
LLM-based agents, characterized by iterative perception, reasoning, and action,
have become a dominant paradigm for complex task automation.
The ReAct~\cite{yao2023react} framework structures agent behavior into interleaved reasoning traces
and tool-invocation actions, grounding language model outputs in external observations.
Multi-agent platforms such as AutoGen~\cite{wu2024autogen} and MetaGPT~\cite{hong2024metagpt}
extend this to collaborative pipelines where specialized agents with distinct roles
coordinate to solve tasks beyond the capacity of a single model.

Tool-augmented reasoning is particularly critical for tasks requiring grounded computation.
Toolformer~\cite{schick2023toolformer} demonstrates that LLMs can autonomously learn
to invoke external APIs (calculators, search engines) at appropriate locations.
In multi-agent data synthesis, MATRIX~\cite{tang-etal-2025-synthesizing} and AgenticMath~\cite{liu2026agenticmathenhancingllmreasoning}
show that specialized generator--verifier pipelines produce higher-quality training data
than single-model generation.

FormalAnalyticGeo adopts this multi-component paradigm with a domain-specific quality layer.
Unlike general-purpose frameworks, our framework embeds a dedicated Quality Verifier
equipped with geometry-specific tools including CDL syntax validation, symbolic solvability checking,
and cross-component answer verification, applied at three gate positions.
This design provides closed-loop feedback that drives upstream retries
until each generated problem meets measurability and correctness thresholds,
yielding annotation-free ground truth for multimodal analytic geometry.

\begin{figure*}[h]
    \centering
    \includegraphics[width=1\linewidth]{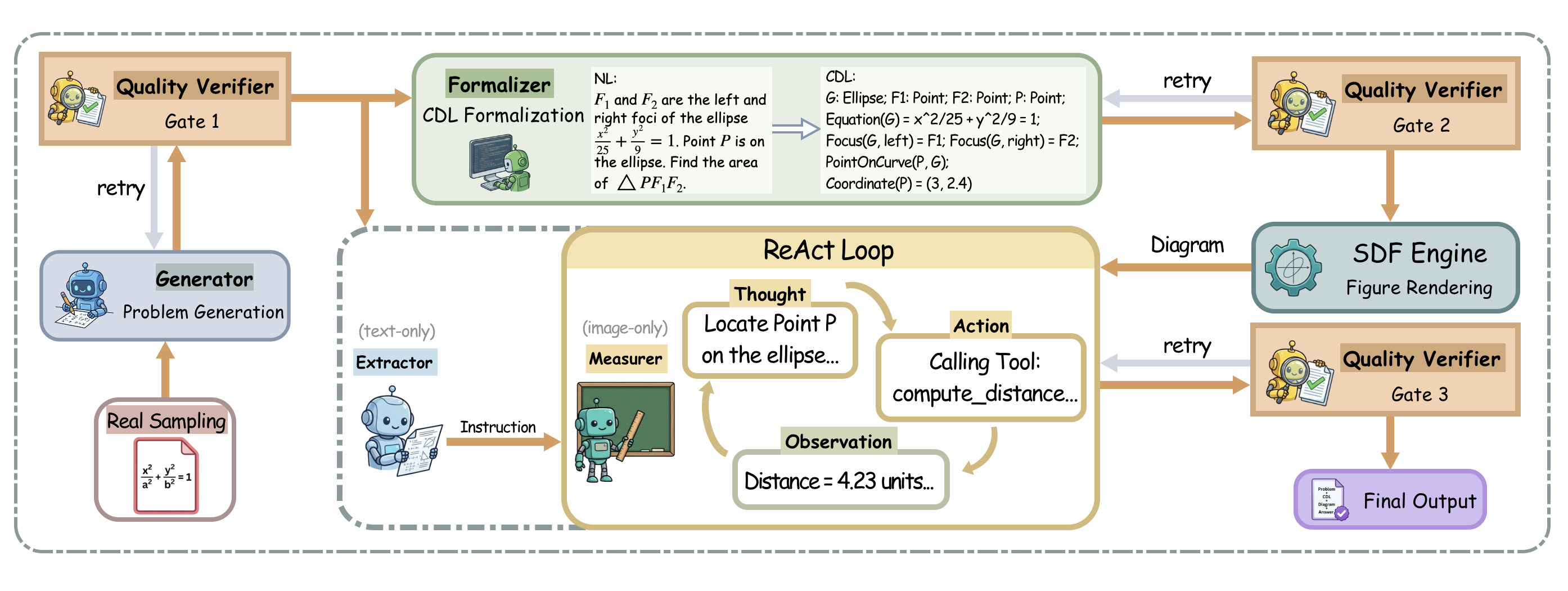}
    \caption{The framework comprises three specialized generative components and one dedicated verification component. Each generative component operates in a task-specific ReAct loop and accepts structured feedback from the Quality Verifier for retry; the Quality Verifier enforces data quality at three gate positions between stages.}
    \label{fig:pipeline}
\end{figure*}

\section{FormalAnalyticGeo}
\label{sec:3}


FormalAnalyticGeo is a scalable framework that automatically generates multimodal analytic geometry problems with formally verified text--diagram consistency. Unlike prior data-generation approaches that rely on template-based synthesis or single-model bootstrapping, FormalAnalyticGeo decomposes the task into specialized stages connected by a formal intermediate language and enforced by quality gates with feedback-driven retry. We first give an overview of the framework, then describe Condition Description Language, a formal language that bridges natural language and diagram rendering, and the SDF-based rendering engine. Finally, we detail the LLM components and quality gates that drive the framework end-to-end.

\subsection{Framework Overview}
\label{sec:3.1}
FormalAnalyticGeo is organized as a sequential system in which four specialized LLM components and one deterministic rendering engine collaborate to produce multimodal analytic geometry problems. Each problem comprises four aligned artifacts: a natural language problem text, a formal CDL annotation, a standardized geometric diagram, and a ground-truth answer. The framework proceeds as follows: the Generator produces a problem together with its conic equation, the Formalizer translates the problem into a CDL program (Section~\ref{sec:3.2}), the SDF engine renders a diagram from the CDL (Section~\ref{sec:3.3}), and the Measurer extracts the ground-truth answer by measuring the rendered diagram.

A key design principle is \emph{closed-loop quality control}. A dedicated Quality Verifier checks each component's output at three gate positions using task-specific verification tools. When a check fails, the Quality Verifier returns structured feedback that pinpoints the error and suggests a correction; the upstream component then retries with this feedback as additional context. This feedback-driven retry mechanism is critical, as unverified outputs can actively degrade downstream data quality. Problems that exhaust all retries are discarded, ensuring that only verified data enters the final dataset.

All inter-agent communication passes through structured formats (CDL programs, measurement instructions, verification verdicts), and the entire framework runs without human intervention.

\subsection{Condition Description Language}
\label{sec:3.2}
The framework requires a formal intermediate representation that is both human-readable and machine-renderable. Existing formal languages for geometry, such as Assertional Logic~\cite{wu-etal-2023-conic10k}, were designed for semantic parsing rather than diagram synthesis and lack a direct mapping to visual primitives. We propose CDL (Condition Description Language), a declarative formal language in which every statement maps to either a visual element or a differentiable constraint in the rendering engine. Renderability is the primary design goal: every valid CDL program can be deterministically compiled into an SDF scene.

\textbf{Statement Categories.} A CDL program is a sequence of semicolon-separated statements organized into seven categories. \emph{Declarations} and \emph{Equations} define geometric objects and their algebraic equations, each creating an SDF primitive with determined parameters. \emph{Coordinates} fix point positions as non-learnable constants. \emph{Derived objects} such as \texttt{Focus}, \texttt{Intersection}, and \texttt{Directrix} compute new entities from existing ones via conic-section formulas at parse time. \emph{Predicates} and \emph{Metric constraints} assert geometric relationships (e.g., \texttt{PointOnCurve}, \texttt{IsTangent}) and numeric measurements (e.g., \texttt{Distance}, \texttt{Slope}), each mapping to a differentiable loss function for the SDF optimizer (Section~3.3). \emph{Inequality constraints} restrict parameter ranges for well-posedness.

\textbf{Machine Verifiability.} CDL's formal structure enables two levels of automatic quality control: syntactic validation by the CDL parser, which detects undeclared names and parsing failures with immediate feedback for the Formalizer's self-correction, and completeness checking by the Quality Verifier, which verifies that all geometric objects in the problem text are encoded in the CDL program (Section~3.4).

\textbf{Scope and Limitations.} The current CDL specification covers five primitive types (line, circle, ellipse, parabola, hyperbola), 12 derived-object operators, and metric and predicate constraints sufficient for the five goal types in our dataset. Region-based rendering (e.g., shading the area inside a curve and above a line) is not yet supported, as it requires signed-region SDF composition beyond point-level constraints; we treat this as future work.

\subsection{SDF Engine}
\label{sec:3.3}
We now describe the rendering engine that compiles CDL programs into geometric diagrams. Traditional plotting libraries such as Matplotlib and TikZ can render curves at given coordinates but treat constraint solving as an external responsibility: when a CDL program specifies that ``point $P$ lies on ellipse $C$ closest to focus $F$,'' the user must resolve $P$'s coordinates separately before plotting. We instead adopt Signed Distance Fields (SDFs) as both the geometric representation and the rendering primitive, unifying constraint solving and rendering in a single differentiable framework. Because SDFs are differentiable, under-determined point positions can be resolved automatically via gradient descent over geometric constraints; because rendering reduces to thresholding on a pixel grid, the resulting coordinate-to-pixel mapping is exact -- a property the Measurer later exploits for visual ground-truth extraction; and because all geometric elements share the same SDF representation, conic curves, coordinate axes, directrix lines, and asymptotes are rendered through one unified mechanism without element-specific drawing logic.

Formally, an SDF is a scalar function $F: \mathbb{R}^2 \to \mathbb{R}$ that assigns to every point $\mathbf{x}$ the signed distance to the nearest boundary of a geometric shape:
\begin{equation}
F(\mathbf{x}) = \begin{cases}
    -d(\mathbf{x}, \partial\Omega) & \text{if } \mathbf{x} \in \Omega, \\
    \phantom{-}d(\mathbf{x}, \partial\Omega) & \text{if } \mathbf{x} \notin \Omega,
\end{cases}
\end{equation}
where $d(\mathbf{x}, \partial\Omega) = \inf_{\mathbf{y} \in \partial\Omega} \|\mathbf{x} - \mathbf{y}\|$. The zero-level set $\{\mathbf{x} : F(\mathbf{x}) = 0\}$ recovers the shape boundary exactly.

\textbf{SDF Primitives.} We implement SDF primitives for the five geometric element types in analytic geometry: line, circle, ellipse, parabola, and hyperbola. Lines and circles admit standard closed-form SDFs. For ellipses and parabolas, we adopt analytical cubic solvers that reduce the closest-point problem to a depressed cubic solvable via Cardano's formula. For hyperbolas, where no closed-form closest-point solution exists, we use the implicit approximation $\text{SDF}_{\text{hyp}}(\mathbf{x}) \approx {|f(\mathbf{x})|}/{\|\nabla f(\mathbf{x})\|}$, where $f(\mathbf{x}) = x^2/a^2 - y^2/b^2 - 1$ is the implicit equation and $\nabla f$ its gradient; this first-order approximation provides sub-pixel accuracy near the curve and converges to the exact SDF on the zero-level set. All primitives are implemented as PyTorch modules with batch dimensions for GPU-accelerated parallel optimization.

\textbf{Constraints and Optimization.} Geometric constraints from the CDL program are compiled into differentiable loss functions over the set of geometric elements $E = \{e_1, e_2, \ldots, e_N\}$. Each constraint $c_i$ maps a configuration to a non-negative scalar (zero when satisfied), and the optimizer seeks the configuration $E^*$ that minimizes the total loss:
\begin{equation}
L(E) = \sum_{i=1}^{m} c_i(E) + \lambda \sum_{j < k} \bigl[\max\bigl(0,\; \tau - \|e_j - e_k\|\bigr)\bigr]^2,
\end{equation}
where the second term is a crowd regularization penalty that prevents elements from collapsing into degenerate configurations by penalizing pairs closer than a threshold~$\tau$. We optimize $L$ with AdamW and cosine-annealing learning rate scheduling over a batch of $B$ parallel initializations, selecting the sample with the lowest loss. For fully determined problems where all coordinates are resolved symbolically from the CDL, optimization is skipped entirely.

\textbf{Rendering} The boundary of each shape is extracted by sampling the SDF on an $N \times N$ pixel grid and thresholding at the zero-level set. All geometric elements -- conic curves, coordinate axes, directrix lines, and asymptotes -- are rendered through this unified SDF mechanism, ensuring pixel-level geometric consistency across all element types. The viewport metadata (coordinate bounds, pixel size) is recorded alongside each image as a sidecar file, providing exact coordinate-to-pixel mappings that the Measurer later uses for visual measurement.


\subsection{Generator}
\label{sec:generator}
Given target objects (line, circle, ellipse, hyperbola, or parabola), the Generator produces a natural language problem together with their polynomial equations. It operates in a ReAct loop with four domain-specific tools: a seed retrieval tool that samples real-world raw text from the Conic10K~\cite{wu-etal-2023-conic10k} dataset, a property enumeration tool that lists the geometric properties derivable from a given equation via SymPy, and two validation tools (rule-based and symbolic) that verify the generated problem's self-consistency. The Generator supports two generation paths: a \emph{forward} path in which it freely composes a problem, and a \emph{reverse} path in which it grounds the problem in symbolically verified properties from the seed library. The reverse path is essential because LLMs often generate problems with incorrect geometric relationships when composing freely; anchoring the generation in SymPy-verified properties significantly reduces such errors. Each problem also carries an explicit information partition: let $I_t$ denote the conditions stated in the problem text and $I_v$ those that must be read from the diagram. The constraint $I_v \cap I_t = \emptyset$ ensures that solving the problem requires both modalities.

\subsection{Formalizer}
\label{sec:Formalizer}
The Formalizer translates the natural language problem into a CDL program at temperature~0. A key design is \emph{grammar prompting}: the CDL specification provided to the Formalizer is dynamically tailored to the detected curve type, pruning irrelevant operators (e.g., removing \texttt{Asymptote} for ellipses) and supplying curve-specific few-shot examples. This reduces attention dilution over the full operator set and prevents the Formalizer from hallucinating unsupported constructs. The output CDL is immediately parsed by the CDL parser; any syntactic error triggers an automatic retry with the parser's error message as feedback, before the result reaches the Quality Verifier.

\subsection{Measurer ReAct Loop}
\label{Measurer}
The Measurer extracts ground-truth answers by measuring the rendered diagram. A critical challenge is \emph{information leakage}: if the Measurer receives both the diagram and the curve equations, the underlying LLM bypasses visual measurement and solves the problem analytically, producing answers that are not true visual ground truth. We address this through \emph{physical information isolation} by splitting the Measurer into two stages. The Task Extractor reads only the problem text and outputs a structured measurement instruction -- specifying what to measure and which entities to locate -- that contains no equations or coordinates. The Visual Measurer receives only this instruction and the diagram image; it never sees the problem text or equations. The Visual Measurer operates in a ReAct loop with a suite of computer-vision tools that cover coordinate mapping, distance and angle measurement, curve probing, and intersection detection. Because we rendered the diagram ourselves, the viewport metadata (coordinate bounds, pixel-to-coordinate mapping) is known exactly, eliminating the axis-reading and scale-detection errors common in diagram understanding.

\subsection{Quality Gates} 
\textbf{Quality Verifier.} A dedicated Quality Verifier enforces end-to-end data quality by independently verifying each stage's output at three gate positions using task-specific tools. On failure, it returns structured feedback that pinpoints the error and suggests a correction; the upstream component retries (up to two additional attempts) with this feedback appended to its context, transforming quality control from a passive filter into an active closed-loop correction process. The three gates are:
\begin{itemize}
\item \textbf{Gate~1} (after problem generation): verifies solvability by attempting a symbolic solution and confirms a finite answer exists.
\item \textbf{Gate~2} (after CDL annotation): validates CDL syntax and checks that all geometric objects mentioned in the problem text are encoded in the CDL.
\item \textbf{Gate~3} (after visual measurement): checks the answer range for plausibility and cross-validates against a text-only symbolic solution.
\end{itemize}
Problems that fail all three attempts at any gate are discarded, ensuring that only verified data enters the final dataset.

\section{AnalyticGeo7k}

We apply the framework described in Section~\ref{sec:3} to generate the AnalyticGeo7k dataset. This section reports dataset statistics, quality, and diversity.

\begin{figure*}
    \centering
    \includegraphics[width=1\linewidth]{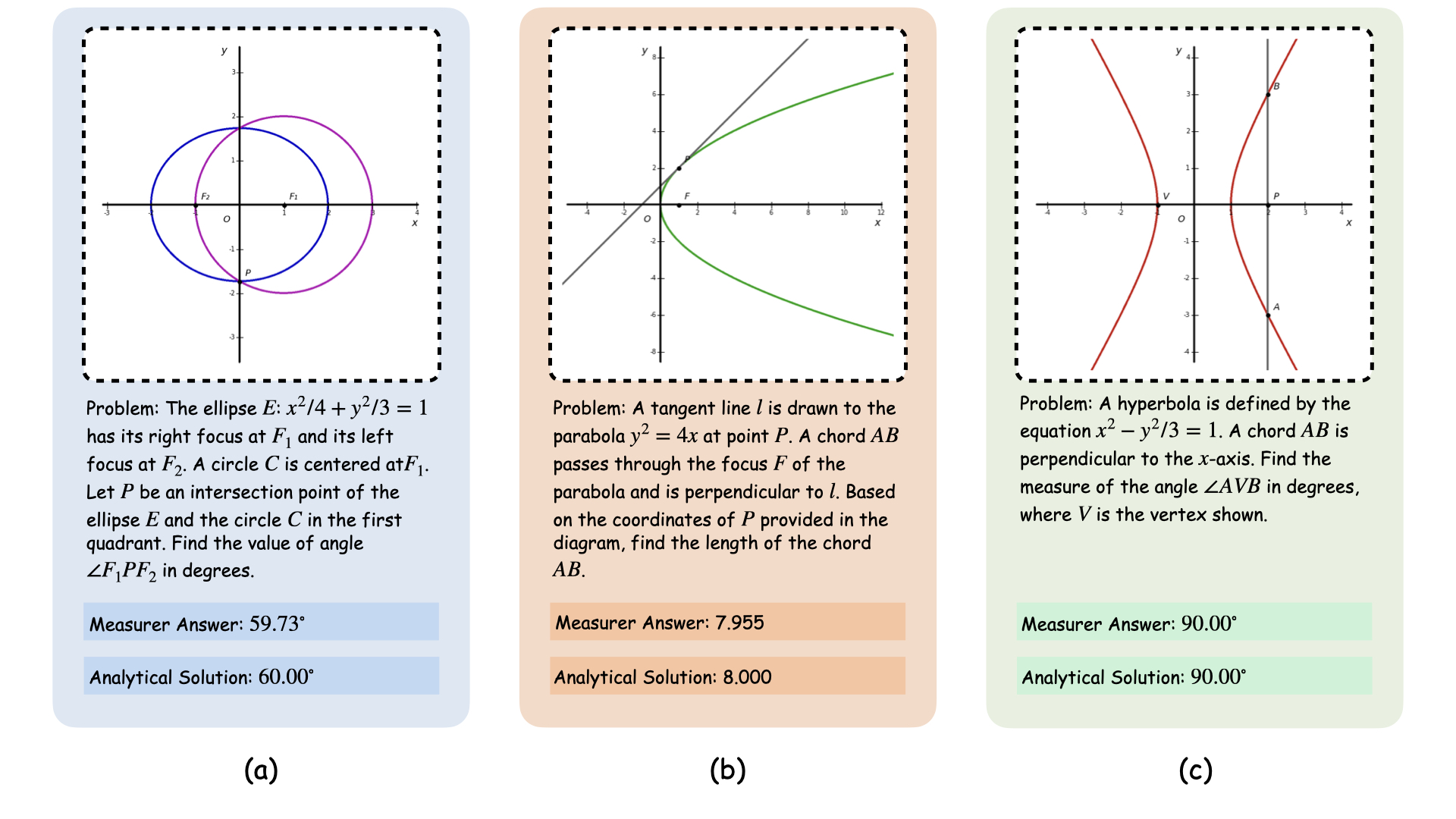}
    \caption{Three representative problems generated by FormalAnalyticGeo, with Measurer answers and analytical solutions.}
    \label{fig:cases}
\end{figure*}


\subsection{Generation Setup}

All four components (Generator, Formalizer, Measurer, and Quality Verifier) use Gemini-3.1-Pro-Preview via an OpenAI-compatible API endpoint. The framework runs in streaming mode: each problem is generated, annotated, rendered, measured, and verified end-to-end before the next problem begins. Target curve types cycle uniformly among ellipse, hyperbola, parabola, and circle; goal types cycle among chord length, area, distance, angle, and perimeter. 

The framework produced a total of 7,823 candidate problems. Of these, 780 were discarded because they exhausted all quality-gate retries without passing. This was typically due to persistent CDL annotation errors or rendering failures, leaving a final dataset of 7,043 verified problems. Each problem takes ${\sim}$5 minutes to generate on average. As the framework is fully automatic and requires no human annotation, the dataset can be scaled to any target size by running additional generation cycles; the 7,043-problem release reported here represents an initial demonstration of the framework's capability.

\subsection{Dataset Statistics}

Table~\ref{tab:cross} summarizes the dataset composition. The 7,043 problems span four curve types and five goal types.
Each curve type contributes 18–32\% of the total, ensuring broad coverage across conic sections. Area and chord length are the most frequent goal types, while distance, angle, and perimeter provide complementary geometric reasoning challenges. 

\begin{table}[H]
\centering
\caption{Curve type $\times$ goal type distribution (number of problems).}
\label{tab:cross}
\resizebox{\columnwidth}{!}{%
\begin{tabular}{l c c c c c | c}
\toprule
 & \textbf{Area} & \textbf{Chord} & \textbf{Dist.} & \textbf{Angle} & \textbf{Perim.} & \textbf{Total} \\
\midrule
Circle    & 485 & 537 & 331 & 465 & 436 & 2{,}254 \\
Ellipse   & 558 & 203 & 406 & 221 & 302 & 1{,}690 \\
Hyperbola & 492 & 428 & 219 & 352 & 340 & 1{,}831 \\
Parabola  & 248 & 307 & 279 & 139 & 295 & 1{,}268 \\
\midrule
\textbf{Total} & 1{,}783 & 1{,}475 & 1{,}235 & 1{,}177 & 1{,}373 & 7{,}043 \\
\bottomrule
\end{tabular}%
}
\end{table}

\subsection{Quality Analysis}

We evaluate dataset quality along two dimensions: framework reliability and ground-truth accuracy.

\textbf{Framework Reliability.} Table~\ref{tab:gate-stats} reports pass rates and retry statistics at each quality gate. \textbf{All three gates achieve $\geq$99\% final pass rate.} Gate~1 reaches 99.1\% on the first try, indicating that the Generator's built-in validation tools effectively prevent unsolvable problems from entering the framework. Gate~2 achieves 98.2\% first-try, with the CDL parser's syntactic auto-repair resolving most issues before the Quality Verifier is invoked; only 1.8\% of problems require a Quality Verifier retry. \textbf{Gate~3 exhibits the highest retry rate (4.9\%)}, reflecting the inherent difficulty of cross-validating visual measurements against symbolic solutions. Problems that exhaust all retries are discarded and replaced.

\begin{table}[H]
\centering
\caption{Quality gate pass rates and retry statistics.}
\label{tab:gate-stats}
\resizebox{\columnwidth}{!}{%
\begin{tabular}{l c c}
\toprule
\textbf{Gate} & \textbf{First-Try Pass} & \textbf{Retry Needed} \\
\midrule
Gate~1 (Problem)  & 99.1\% & 0.9\%  \\
Gate~2 (CDL)      & 98.2\% & 1.8\%  \\
Gate~3 (Answer)   & 95.1\% & 4.9\%  \\
\midrule
Formalizer L1 syntax repair & 86.6\% & 13.4\% \\
\bottomrule
\end{tabular}%
}
\end{table}

\textbf{Ground-Truth Accuracy.} To evaluate the accuracy of the Measurer's visual measurements, we sample 164 problems from the dataset (balanced across curve and goal types) and manually compute their exact symbolic answers. For each sampled problem, we derive the analytical solution using standard analytic geometry techniques and compute the relative error $\epsilon = |a_{\text{meas}} - a_{\text{exact}}| / |a_{\text{exact}}|$.

\begin{figure}[h]
    \centering
    \includegraphics[width=0.75\linewidth]{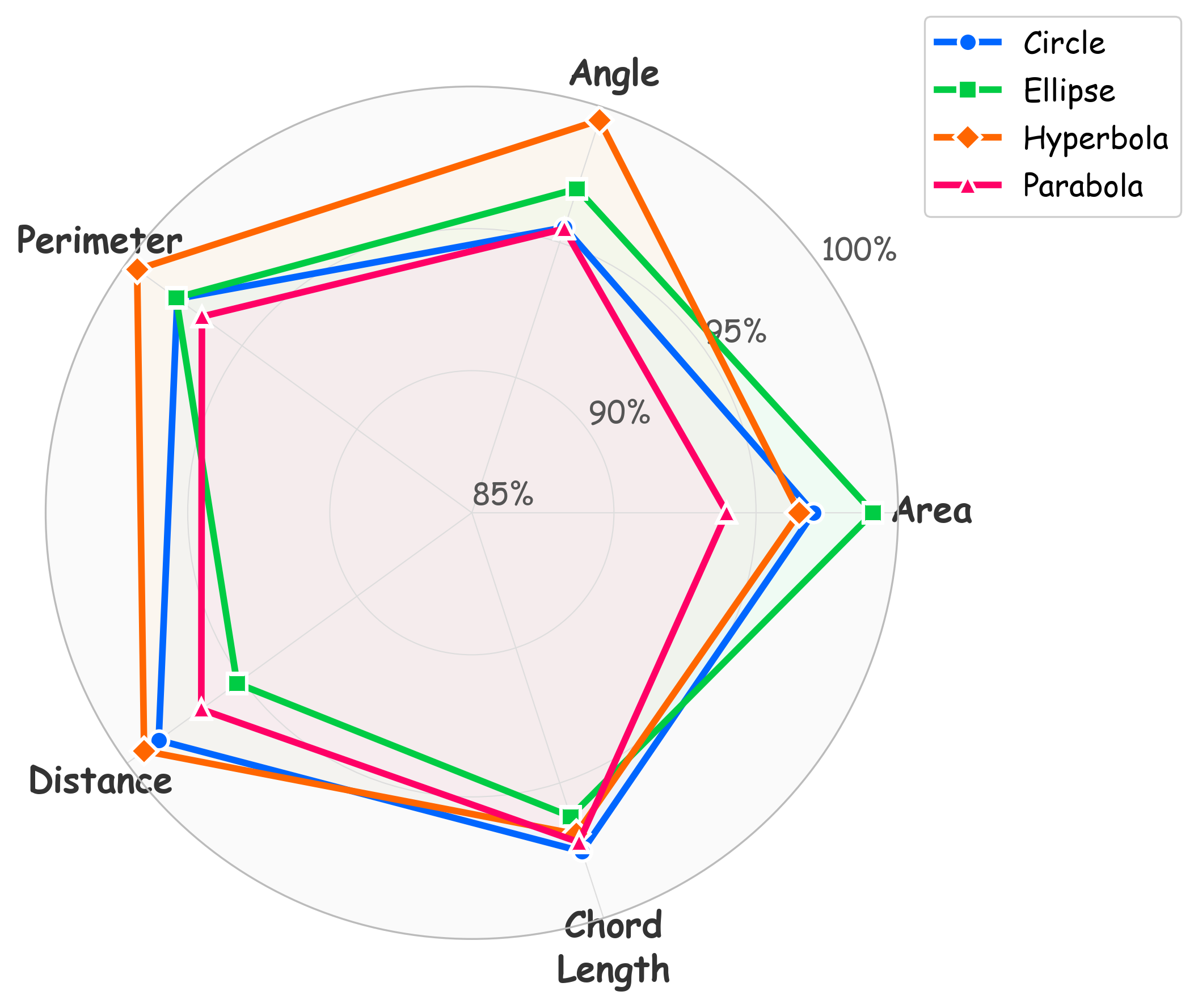}
    \caption{Per-category ground-truth accuracy ($1 - \bar{\epsilon}$, zoomed to 85--100\%). Parabola--area is the hardest combination (93.97\%).}
    \label{fig:radar}
\end{figure}

\textbf{Across the 164 sampled problems, the Measurer achieves a median relative error of 0.70\%} (mean 2.84\%), with 82.3\% of answers falling within 5\% of the exact solution. Figure~\ref{fig:radar} shows the accuracy breakdown by curve type and goal type. Most categories achieve $>$95\% accuracy (measured as $1 - \bar{\epsilon}$). \textbf{Parabola--area problems are the hardest category} (93.97\% accuracy). We attribute this to multi-step visual reasoning: measuring a parabolic segment area requires identifying the tangent line, computing intercepts, and deriving the enclosed area, with each step accumulating measurement error from successive tool calls. This category-specific pattern suggests a fundamental trade-off in visual ground-truth extraction: single-step measurements (distances, angles) are highly accurate, while multi-step derivations amplify per-step errors. Mitigating this accumulation is a direction for future work; potential approaches include intermediate result verification between tool calls and symbolic cross-checking of partial measurements before combining them into a final answer.
The 1\% evaluation threshold in Table~\ref{tab:baseline} remains reliable under this noise level, as the median GT error (0.70\%) is well below the threshold.

\section{Experiment}
\label{sec:experiment}

\subsection{Experimental Results}
\label{sec:exprst}
We evaluate eight representative MLLMs on the AnalyticGeo7K benchmark under both image and text-only input modes. Table~\ref{tab:baseline} reports accuracy broken down by five goal types and four curve types.

\textbf{Challenging Nature of Analytic Geometry.} The best-performing model, Gemini 3 Flash with image input, achieves an overall accuracy of 77.6\%, followed by GPT-5.2 (73.1\%) and Claude Opus 4.5 (72.0\%). GPT-4o reaches only 23.2\%, indicating that analytic geometry problems remain difficult even for frontier MLLMs. Performance drops sharply in text-only mode: the best text-only result (GPT-5.2, 42.0\%) trails the best image result by over 35 percentage points, and weaker models such as Mistral Large (18.4\%) and GPT-4o (18.8\%) fall below 20\%. This substantial gap suggests that jointly reasoning about coordinate systems, algebraic curves, and geometric relationships poses a significant challenge to current models. Moreover, no single model dominates all fine-grained categories: Claude Opus 4.5 leads on Area (79.4\%) and Angle (88.0\%) but ranks only third on Chord, while GPT-5.2 leads on Distance (81.3\%) but falls behind on Area (68.0\%), indicating complementary strengths across goal types and curve families.

\textbf{Impact of Visual Input.} Image-input accuracy substantially exceeds text-only performance for all models, confirming that diagrams provide informative visual cues for analytic geometry reasoning. Gemini 3 Flash exhibits the largest gap (77.6\% $\to$ 36.0\%, $-$41.6\,pp), while GPT-5.2 shows a smaller but still significant drop (73.1\% $\to$ 42.0\%, $-$31.1\,pp). The gap is especially pronounced on specific curve types: Gemini's Circle accuracy drops from 85.3\% to 33.0\% ($-$52.3\,pp) without images. Interestingly, the text-only ranking partially reshuffles: GPT-5.2 overtakes Gemini 3 Flash (42.0\% vs.\ 36.0\%), suggesting that some models rely more heavily on visual parsing while others maintain stronger algebraic reasoning in the absence of diagrams.

\begin{table*}[!t]
\centering
\caption{Model performance on AnalyticGeo7K benchmark (\%). Accuracy = answer within 1\% relative error of ground truth. The \textcolor{red}{\textbf{best}} and \textcolor{blue}{\underline{second best}} results per section are highlighted.}
\label{tab:baseline}
\resizebox{\textwidth}{!}{%
\begin{tabular}{l>{\columncolor{gray!15}}c|ccccc|cccc}
\toprule
 & & \multicolumn{5}{c|}{\textbf{Goal Type}} & \multicolumn{4}{c}{\textbf{Curve Type}} \\
\cmidrule(lr){3-7} \cmidrule(lr){8-11}
\textbf{Model} & \textbf{Overall} & \textbf{Area} & \textbf{Chord} & \textbf{Dist.} & \textbf{Angle} & \textbf{Perim.} & \textbf{Circle} & \textbf{Ellipse} & \textbf{Hyper.} & \textbf{Parab.} \\
\midrule
\rowcolor{orange!10}
\multicolumn{11}{l}{\textit{Image Input}} \\
GPT-4o~\cite{openai2024gpt4ocard}            &  23.2 & 23.8 & 14.3 & 24.1 & 37.0 & 19.4 & 24.5 & 20.5 & 23.3 & 24.1 \\
Claude Sonnet 4~\cite{anthropic2025claude4}   &  49.4 & 59.5 & 42.9 & 41.4 & 59.3 & 41.9 & 60.4 & 59.0 & 34.9 & 37.9 \\
Qwen-VL-Max~\cite{bai2025qwen25vltechnicalreport}       &  69.0 & 69.6 & 53.9 & 70.3 & 79.9 & 74.9 & 66.5 & \textcolor{blue}{\underline{80.2}} & 65.4 & \textcolor{red}{\textbf{71.9}} \\
Claude Opus 4.5~\cite{anthropic2025opus45}   &  72.0 & \textcolor{red}{\textbf{79.4}} & 60.9 & 68.5 & \textcolor{red}{\textbf{88.0}} & 70.0 & \textcolor{blue}{\underline{79.0}} & \textcolor{red}{\textbf{82.3}} & 61.0 & 62.9 \\
GPT-5.2~\cite{openai2025gpt52}           & \textcolor{blue}{\underline{73.1}} & 68.0 & \textcolor{blue}{\underline{64.1}} & \textcolor{red}{\textbf{81.3}} & 79.8 & \textcolor{blue}{\underline{76.4}} & 72.0 & 76.0 & \textcolor{red}{\textbf{73.4}} & \textcolor{blue}{\underline{70.5}} \\
Gemini 3 Flash~\cite{google2025gemini3flash}    & \textcolor{red}{\textbf{77.6}} & \textcolor{blue}{\underline{74.3}} & \textcolor{red}{\textbf{71.0}} & \textcolor{blue}{\underline{78.7}} & \textcolor{blue}{\underline{85.3}} & \textcolor{red}{\textbf{78.1}} & \textcolor{red}{\textbf{85.3}} & 76.2 & \textcolor{blue}{\underline{71.4}} & 68.3 \\
\midrule
\rowcolor{orange!10}
\multicolumn{11}{l}{\textit{Text Only}} \\
Mistral Large~\cite{mistral2024large}      &  18.4 & 26.1 & 12.2 & 16.3 & 12.5 & 24.2 & 18.6 & 24.5 & 11.5 & 19.6 \\
GPT-4o~\cite{openai2024gpt4ocard}            &  18.8 & 26.7 & 12.5 & 16.7 &  6.2 & 30.8 & 19.0 & 25.0 & 11.8 & 20.0 \\
Claude Sonnet 4~\cite{anthropic2025claude4}   &  25.0 & 32.5 & 12.1 & 17.3 & 39.2 & 24.4 & 28.3 & 39.0 & 17.5 & 14.1 \\
DeepSeek-V3.1~\cite{deepseekai2025deepseekv3technicalreport}      &  30.9 & 33.9 & 14.2 & 29.6 & 54.2 & 32.5 & 30.9 & 43.6 & 23.8 & 26.2 \\
Qwen-VL-Max~\cite{bai2025qwen25vltechnicalreport}       &  35.0 & 31.0 & 14.0 & \textcolor{red}{\textbf{38.2}} & \textcolor{red}{\textbf{64.7}} & 35.1 & 32.7 & 46.8 & 28.2 & \textcolor{red}{\textbf{34.6}} \\
Gemini 3 Flash~\cite{google2025gemini3flash}    &  36.0 & 43.7 & \textcolor{blue}{\underline{19.4}} & 24.0 & 41.4 & \textcolor{red}{\textbf{50.9}} & 33.0 & 49.1 & \textcolor{red}{\textbf{41.2}} & 15.9 \\
Claude Opus 4.5~\cite{anthropic2025opus45}   & \textcolor{blue}{\underline{38.0}} & \textcolor{red}{\textbf{46.0}} & 18.0 & 29.9 & 57.9 & 42.5 & \textcolor{blue}{\underline{38.7}} & \textcolor{red}{\textbf{58.1}} & 32.0 & 24.4 \\
GPT-5.2~\cite{openai2025gpt52}           & \textcolor{red}{\textbf{42.0}} & \textcolor{blue}{\underline{44.6}} & \textcolor{red}{\textbf{20.0}} & \textcolor{blue}{\underline{35.8}} & \textcolor{blue}{\underline{62.8}} & \textcolor{blue}{\underline{50.7}} & \textcolor{red}{\textbf{40.4}} & \textcolor{blue}{\underline{57.0}} & \textcolor{blue}{\underline{39.3}} & \textcolor{blue}{\underline{29.0}} \\
\bottomrule
\end{tabular}
}
\end{table*}

\subsection{Ablation Study}
\label{sec:ablation}

We study four design decisions in FormalAnalyticGeo,
each isolating one framework component (Tables~\ref{tab:ablation} and \ref{tab:fewshot}).
\textit{w/o Quality Verifier} disables all three quality gates and feedback-driven retry loops,
while retaining each component's internal self-validation tools,
so that this variant isolates the contribution of cross-component external verification.
\textit{w/o Seed Library} removes the \textit{seed\_pool} tool from the Generator,
eliminating in-context exposure to verified seed problems;
because the reverse path accounts for only 3.7\% of baseline outputs,
the primary effect is the loss of implicit structural priors for forward-path generation.
\textit{Formalizer Few-shot Gradient} re-runs only the Formalizer on 200 existing problems
at four few-shot counts ($N \in \{0, 4, 8, 10\}$) to measure CDL annotation quality
without re-generating problems or diagrams.
\textit{w/o CV Tools} removes all measurement tools from the Visual Measurer,
forcing visual parameter estimation from the rendered diagram alone;
this variant requires no new data collection
since it is evaluated on 200 problems sampled from the 7,043-problem dataset.

\begin{table}[H]
\centering
\caption{Ablation study results (200 problems per variant).}
\label{tab:ablation}
\resizebox{\columnwidth}{!}{%
\begin{tabular}{lccc}
\toprule
\textbf{Variant} & \textbf{G3 retry (\%)} & \textbf{Within 5\% (\%)} & \textbf{Mean err.\ (\%)} \\
\midrule
Full framework       & 4.9  & 82.3 & 2.84  \\
\midrule
w/o Quality Verifier & ---  & 45.5 & 35.7  \\
w/o Seed Library     & 26.5 & 72.9 & 9.92  \\
w/o CV Tools         & ---  & 49.0 & 40.5  \\
\bottomrule
\end{tabular}%
}
\end{table}

\begin{table}[H]
\centering
\caption{Formalizer few-shot gradient (200 problems, balanced curve types).}
\label{tab:fewshot}
\begin{tabular}{ccc}
\toprule
\textbf{Few-shot $N$} & \textbf{L1 first-try (\%)} & \textbf{Avg.\ attempts} \\
\midrule
0 (zero-shot) & 60.0 & 1.58 \\
4             & 62.0 & 1.48 \\
8             & 76.0 & 1.34 \\
10 (baseline) & 84.0 & 1.18 \\
\bottomrule
\end{tabular}
\end{table}

Removing all Quality Verifier gates causes a dramatic accuracy drop:
within-5\% accuracy falls from 82.3\% to 45.5\%
and mean relative error rises from 2.84\% to 35.7\%.
CDL completeness remains 100\% under human review,
indicating the degradation stems not from missing problem structure
but from compounded visual measurement errors in the Visual Measurer
that the gate's \texttt{solve\_text\_only} verification would have caught and triggered retries for.
This confirms that cross-component external verification provides a substantial quality lift
over each component's self-validation alone.

Removing seed library access from the Generator increases the Gate~3 retry rate
from 4.9\% to 26.5\%, reflecting that forward-path problems are geometrically less well-formed
without in-context structural priors from verified examples.
Within-5\% accuracy drops 9.4 percentage points to 72.9\%
and mean error rises to 9.92\%,
despite Gate~1 and Gate~2 pass rates remaining unaffected,
confirming that the seed library's contribution lies not in enabling generation
but in improving the geometric coherence of generated problems.

Without CV tools, the Visual Measurer is forced to visually estimate geometric parameters
and apply analytical formulas, compounding parameter-reading errors into the final answer:
a small misread of an ellipse's semi-axis $a$, for example,
propagates through the chord-length formula to yield errors of 30--80\%.
Within-5\% accuracy drops to 49.0\% and mean relative error rises to 40.5\%,
a 33.3 percentage point decline that directly quantifies
the contribution of structured pixel-level measurement to GT extraction precision.

Table~\ref{tab:fewshot} shows a monotone positive trend:
L1 first-try pass rate rises from 60.0\% to 84.0\% as $N$ increases from 0 to 10,
and average attempts decrease from 1.58 to 1.18.
The largest single gain (+14~pp) occurs between $N=4$ and $N=8$,
where most curve types first receive at least three curve-specific examples;
below this threshold, certain types have only one example,
insufficient for the Formalizer to infer CDL's predicate names and argument ordering.

\section{Conclusion}

We presented FormalAnalyticGeo, a scalable framework that generates multimodal analytic geometry problems without human annotation. By unifying CDL as a formal intermediate representation, SDF-based rendering for geometrically exact diagrams, and a four-component generation loop supervised by a Quality Verifier with feedback-driven retries, the framework produces over 7K verified problems with a median ground-truth error of 0.70\%. Ablation experiments confirm that each component contributes measurably: removing the Quality Verifier drops within-5\% accuracy from 82.3\% to 45.5\%, and disabling CV tools raises mean error from 2.84\% to 40.5\%. The CDL specification and SDF engine are designed to be extensible; broader analytic geometry categories such as locus problems and parametric equations can be incorporated by adding corresponding primitives without modifying the core framework.


\bibliographystyle{ACM-Reference-Format}
\bibliography{sample-base}


\newpage

\appendix

\section{CDL Specification}
\label{app:cdl}

This appendix provides the complete operator catalog of the Condition Description Language (CDL) introduced in Section~\ref{sec:3.2}.
A CDL program is a sequence of semicolon-separated statements that collectively specify an analytic-geometry diagram suitable for rendering.

\paragraph{Declarations and Equations.}
Every geometric object must be declared with a name and a type before use.
Supported types include \texttt{Point}, \texttt{Origin} (fixed at the coordinate origin $(0,0)$), \texttt{Line}, \texttt{Circle}, \texttt{Ellipse}, \texttt{Hyperbola}, \texttt{Parabola}, and \texttt{Number} (an algebraic parameter).
An \texttt{Equation} statement binds an algebraic equation to a declared curve or line; the parser classifies the equation by discriminant analysis of the general quadratic form $Ax^2 + Bxy + Cy^2 + Dx + Ey + F = 0$ and extracts the curve parameters automatically.
A \texttt{Coordinate} statement fixes a point at a specified position.

Table~\ref{tab:cdl_operators} lists all 35 operators, organized into three categories: derived-object operators, predicates, and metric constraints.

\begin{table}[h]
\centering
\caption{Complete CDL operator catalog: 12 derived-object operators, 8 predicates, and 15 metric constraints.}
\label{tab:cdl_operators}
\resizebox{\columnwidth}{!}{%
\begin{tabular}{@{}ll@{}}
\toprule
\rowcolor{orange!10}
\multicolumn{2}{@{}l}{\textbf{Derived-Object Operators} --- \emph{compute new entities from existing ones}} \\
\texttt{Focus}(curve, [dir]) $\to$ Point           & \texttt{Vertex}(curve, [dir]) $\to$ Point \\
\texttt{Center}(curve) $\to$ Point                  & \texttt{Directrix}(curve, [dir]) $\to$ Line \\
\texttt{Asymptote}(hyp.) $\to$ Lines               & \texttt{MidPoint}(pt, pt) $\to$ Point \\
\texttt{Intersection}(obj, obj) $\to$ Points        & \texttt{TangentOnPoint}(pt, crv) $\to$ Line \\
\texttt{TangentOfPoint}(pt, crv) $\to$ Lines        & \texttt{TangentPoint}(ln, crv) $\to$ Point \\
\texttt{Projection}(pt, axis/ln) $\to$ Point        & \texttt{FootPoint}(pt, ln) $\to$ Point \\
\midrule
\rowcolor{orange!10}
\multicolumn{2}{@{}l}{\textbf{Predicates} --- \emph{assert geometric relationships (SDF constraint)}} \\
\texttt{PointOnCurve(P, G)}: $P$ on $G$             & \texttt{IsPerpendicular(l1, l2)}: $l_1 \perp l_2$ \\
\texttt{IsParallel(l1, l2)}: $l_1 \parallel l_2$    & \texttt{IsTangent(l, G)}: $l$ tangent to $G$ \\
\texttt{IsChordOf(l, G)}: chord of $G$              & \texttt{IsDiameter(seg, C)}: diameter of $C$ \\
\texttt{IsInTangent(C1, C2)}: int.\ tangent         & \texttt{IsOutTangent(C1, C2)}: ext.\ tangent \\
\midrule
\rowcolor{orange!10}
\multicolumn{2}{@{}l}{\textbf{Metric Constraints} --- \emph{numeric measurements; illustrative examples}} \\
\texttt{Distance(P, F1) = 2}                        & \texttt{Slope(l) = 3} \\
\texttt{Eccentricity(G) = 1/2}                      & \texttt{FocalLength(G) = 6} \\
\texttt{HalfFocalLength(G) = 3}                     & \texttt{Inclination(l) = pi/4} \\
\texttt{Length(SemiMajorAxis(G)) = 3}                & \texttt{Area(TriangleOf(A,O,B)) = 4} \\
\texttt{Perimeter(TriangleOf(A,B,C)) = 12}          & \texttt{Abs(LineSegmentOf(A,B)) = 5} \\
\texttt{AngleOf(P, F1, F2) = 90}                    & \texttt{XCoordinate(P) = 3} \\
\texttt{YCoordinate(P) = -1}                        & \texttt{Quadrant(P) = 1} \\
\multicolumn{2}{@{}l}{\texttt{DotProduct(VectorOf(A,B), VectorOf(C,D)) = 0}} \\
\bottomrule
\end{tabular}}
\end{table}

\paragraph{Derived Objects.}
Derived-object operators compute new geometric entities from existing ones.
Operators that produce points (e.g., \texttt{Focus}, \texttt{MidPoint}) evaluate symbolically at parse time from the parent curve's parameters.
\texttt{Intersection} is resolved analytically by substitution into the curve equation and may return multiple points via set-assignment syntax:

\smallskip
\centerline{\texttt{\{A, B\} = Intersection(l, G)}}
\smallskip

\texttt{Focus}, \texttt{Vertex}, and \texttt{Directrix} accept an optional \texttt{direction} parameter drawn from \{\texttt{left}, \texttt{right}, \texttt{upper}, \texttt{lower}\}.
It may be omitted when the curve admits a unique instance, e.g., the single focus of a parabola.
In Table~\ref{tab:cdl_operators}, square brackets denote optional arguments.

Note that \texttt{Projection} maps a point onto a coordinate axis (or an axis-aligned line), whereas \texttt{FootPoint} computes the perpendicular foot from a point onto an arbitrary line.

CDL distinguishes three tangent-related operators.
\texttt{TangentOnPoint}(P, G) computes the tangent line to curve $G$ at a point $P$ that already lies on $G$, via implicit differentiation of the curve equation.
\texttt{TangentOfPoint}(P, G) finds tangent lines from an external point $P$ to $G$ using discriminant analysis and may yield multiple solutions.
\texttt{TangentPoint}(l, G) returns the contact point(s) where a known tangent line $l$ touches $G$.

\paragraph{Predicates.}
Each predicate asserts a geometric relationship and compiles to a differentiable SDF constraint whose satisfaction corresponds to a zero loss value.
For instance, \texttt{PointOnCurve(P, G)} penalizes the signed distance from $P$ to $G$, while \texttt{IsTangent(l, G)} enforces that the discriminant of the line--curve intersection system vanishes.
\texttt{IsInTangent} and \texttt{IsOutTangent} constrain the distance between two circle centers to equal $|r_1 - r_2|$ or $r_1 + r_2$, respectively.
Predicates carry no explicit numeric target; they are satisfied when the constraint residual reaches zero.

\paragraph{Metric Constraints.}
A metric constraint binds a geometric measurement to a target value, creating a loss term $|f_{\text{measured}} - f_{\text{target}}|$.
The measurement function is evaluated from the current configuration: \texttt{Distance} computes Euclidean distance, \texttt{Area} uses the cross-product formula, and \texttt{AngleOf} applies the law of cosines.
Algebraic parameters declared as \texttt{Number} may appear as targets, enabling parametric constraints such as \texttt{Length(SemiMajorAxis(G)) = a}.

\paragraph{Inequality Constraints and Auxiliary Wrappers.}
Inequality constraints (e.g., \texttt{a > 0}; \texttt{a > b}) restrict the admissible ranges of algebraic parameters.
CDL also provides auxiliary wrappers for composing references inside other operators:

\begin{itemize}
\item \texttt{LineSegmentOf(A,\,B)} --- line segment through $A$ and $B$.
\item \texttt{OverlappingLine(\ldots)} --- the full line through two points.
\item \texttt{RightPart(G)}, \texttt{LeftPart(G)} --- individual branches of a hyperbola.
\item \texttt{TriangleOf(A,\,B,\,C)}, \texttt{VectorOf(A,\,B)} --- point-tuple wrappers for area/perimeter and dot-product metrics.
\end{itemize}

\paragraph{Worked Example.}
The following CDL program encodes the problem:
\emph{``From the left focus $F_1$ of the hyperbola $\tfrac{x^2}{16} - \tfrac{y^2}{25} = 1$, draw a line tangent to the circle $x^2 + y^2 = 16$, touching it at point $T$. Extend $F_1T$ to meet the right branch at $P$, and let $M$ be the midpoint of $F_1P$.''}

\begin{cdlcode}
G: Hyperbola; H: Circle; F1: Point; Z: Line
T: Point; P: Point; M: Point; O: Origin
Equation(G) = x^2/16 - y^2/25 = 1
Equation(H) = x^2 + y^2 = 16
Focus(G, left) = F1
TangentOfPoint(F1, H) = Z
TangentPoint(Z, H) = T
Intersection(OverlappingLine(LineSegmentOf(F1, T)),
             RightPart(G)) = P
MidPoint(LineSegmentOf(F1, P)) = M
\end{cdlcode}

\noindent
This nine-statement program declares four points, one line, and two curves.
It derives the left focus symbolically, computes the tangent line via discriminant analysis, intersects the extended ray with the right branch of the hyperbola, and determines the midpoint~$M$.
The CDL parser compiles these statements into constraint tuples that are consumed by the SDF-based rendering engine.

\section{Signed Distance Field Formulation and Constraint Details}
\label{app:sdf_details}

In this appendix, we provide a comprehensive mathematical formulation of the Signed Distance Field (SDF) representation, the differentiable loss functions for geometric constraints, the optimization procedure, and the boundary extraction method used in our diagram synthesis engine. This material supplements the description in Section~\ref{sec:3.3} of the main paper.

\subsection{SDF Definition and Basic Representations}
\label{app:sdf_def}

A Signed Distance Field (SDF) is a scalar field that assigns to every point \(\mathbf{x} \in \mathbb{R}^2\) the signed distance to the closest point on the boundary of a geometric shape. Formally, for a shape \(\Omega \subset \mathbb{R}^2\) with boundary \(\partial\Omega\), the SDF \(F(\mathbf{x})\) is defined as:
\[
F(\mathbf{x}) = 
\begin{cases}
\displaystyle \min_{\mathbf{y} \in \partial\Omega} \|\mathbf{x} - \mathbf{y}\|, & \text{if } \mathbf{x} \text{ is outside } \Omega, \\[1.2em]
\displaystyle -\min_{\mathbf{y} \in \partial\Omega} \|\mathbf{x} - \mathbf{y}\|, & \text{if } \mathbf{x} \text{ is inside } \Omega,
\end{cases}
\]
where the sign convention follows the common “outside positive, inside negative” rule. Equivalently, the unsigned distance is given by:
\[
d(\mathbf{x}, \partial\Omega) = \inf_{\mathbf{y} \in \partial\Omega} \|\mathbf{x} - \mathbf{y}\|,
\]
and the sign is determined by a point-in-polygon test for polygons or by the side of the curve for lines and circles.

In our framework, each geometric element is represented by its own SDF, which depends on a small set of continuous parameters. Table~\ref{tab:sdf_primitives} lists the SDF expressions for all primitives used in our rendering engine, covering both basic elements and analytic-geometry curves.

\begin{table*}[t]
\centering
\caption{SDF expressions for all geometric primitives in our rendering engine.}
\label{tab:sdf_primitives}
\begin{tabular}{lll}
\toprule
\textbf{Primitive} & \textbf{SDF } \(F(x,y)\) & \textbf{Method} \\
\midrule
\rowcolor{orange!10}
\multicolumn{3}{@{}l}{\textit{Basic Primitives}} \\
Point \(\mathbf{p}=(p_x,p_y)\) & \(\sqrt{(x-p_x)^2 + (y-p_y)^2}\) & Exact \\
Line through \(\mathbf{a},\mathbf{b}\) & \(\dfrac{|(y-a_y)(b_x-a_x) - (x-a_x)(b_y-a_y)|}{\|\mathbf{b}-\mathbf{a}\|}\) & Exact \\
Segment \(\overline{\mathbf{a}\mathbf{b}}\) & \(\|\mathbf{x}-\mathbf{a} - \mathrm{clamp}(t,0,1)(\mathbf{b}-\mathbf{a})\|\), \; $t = \frac{(\mathbf{x}-\mathbf{a})\cdot(\mathbf{b}-\mathbf{a})}{\|\mathbf{b}-\mathbf{a}\|^2}$ & Exact \\
Circle \((\mathbf{c}, r)\) & \(\bigl|\|\mathbf{x}-\mathbf{c}\| - r\bigr|\) & Exact \\
\midrule
\rowcolor{orange!10}
\multicolumn{3}{@{}l}{\textit{Analytic Geometry Curves}} \\
Ellipse \(\tfrac{x^2}{a^2}+\tfrac{y^2}{b^2}=1\) & Closest-point via cubic equation (Cardano / trigonometric branch) & Exact \\
Parabola \(y^2=4px\) & Closest-point via cubic equation; four directions (up/down/left/right) & Exact \\
Hyperbola \(\tfrac{x^2}{a^2}-\tfrac{y^2}{b^2}=1\) & \(\dfrac{|f(\mathbf{x})|}{\|\nabla f(\mathbf{x})\|}\), \; $f = \tfrac{x^2}{a^2}-\tfrac{y^2}{b^2}-1$, \; $\nabla f = (\tfrac{2x}{a^2}, -\tfrac{2y}{b^2})$ & Approx. \\
\bottomrule
\end{tabular}
\end{table*}

The basic primitives (point, line, segment, circle) admit closed-form SDF expressions.
For ellipses and parabolas, we adopt the analytical method of Qu\'{\i}lez: the closest-point problem reduces to a depressed cubic, solved via Cardano's formula when the discriminant $d \ge 0$ or a trigonometric substitution when $d < 0$.
For hyperbolas, an exact SDF would require solving a quartic; we instead use the implicit approximation $|f|/\|\nabla f\|$, which is accurate near the curve boundary where rendering occurs.
All expressions are continuous and differentiable almost everywhere, making them suitable for gradient-based optimization and threshold-based rendering.

\subsection{Geometric Constraint Loss Functions}
\label{app:constraints}

Geometric relationships between elements are encoded as differentiable loss functions that become zero when the constraint is satisfied. Table~\ref{tab:constraint_losses} summarises the key constraints. Each loss is non-negative and smooth.

\begin{table}[h]
\centering
\caption{Differentiable loss functions for geometric constraints used in analytic-geometry rendering.}
\label{tab:constraint_losses}
\begin{tabular}{ll}
\toprule
\textbf{Constraint} & \textbf{Loss Function} \\
\midrule
\textbf{Equality} \(A = B\) & \(|A - B|\) \\
\rowcolor{gray!8}
\textbf{Inequality} \(A \le B\) & \(\max(A - B, 0)\) \\
\textbf{Point on curve} \(P \in G\) & \(F_G(P)\) \\
\rowcolor{gray!8}
\textbf{Slope} of line \(l\) equals $m$ & \(|\mathrm{slope}(l) - m|\) \\
\textbf{Perpendicularity} \(l_1 \perp l_2\) & \(|\mathbf{v}_1 \cdot \mathbf{v}_2|\) \\
\rowcolor{gray!8}
\textbf{Angle} \(\angle ABC\) & \(\displaystyle \arccos\!\left(\frac{\vec{BA}\cdot\vec{BC}}{\|\vec{BA}\|\,\|\vec{BC}\|}\right)\) \\
\textbf{Area} of \(\triangle ABC\) & \(\displaystyle \tfrac{1}{2}|(\mathbf{B}-\mathbf{A}) \times (\mathbf{C}-\mathbf{A})|\) \\
\rowcolor{gray!8}
\textbf{Distance} \(\|PQ\|\) & \(\sqrt{(p_x - q_x)^2 + (p_y - q_y)^2}\) \\
\textbf{Tangency} (line $l$ tangent to $G$) & Discriminant of $l \cap G = 0$ \\
\rowcolor{gray!8}
\textbf{Crowd penalty} & \(\displaystyle \sum_{i<j} \max\bigl(0,\tau - \|\mathbf{x}_i - \mathbf{x}_j\|\bigr)^2\) \\
\bottomrule
\end{tabular}
\end{table}

\noindent
The \emph{point-on-curve} constraint directly uses the SDF value: for a point $P$ that should lie on curve $G$, the loss is $F_G(P)$, which reaches zero exactly on the curve boundary.
The \emph{tangency} constraint substitutes the line equation into the curve equation and penalises the absolute value of the discriminant; when the discriminant vanishes, the line is tangent.
The crowd penalty prevents distinct points from collapsing to the same location during optimisation; it is vectorised for efficient GPU computation.

\subsection{Optimization Algorithm and Hyperparameters}
\label{app:optimization}

In analytic geometry, most geometric quantities are determined analytically from the curve equation: foci, vertices, asymptotes, directrices, and intersection points are all computed symbolically by the CDL parser.
Consequently, the majority of diagrams require \emph{no} gradient-based optimisation---the SDF objects are constructed with fixed parameters and rendered directly.

Optimisation is invoked only when the diagram contains \emph{free points} whose positions are constrained but not uniquely determined by the equations (e.g., a point on a curve satisfying an additional distance constraint).
In such cases, the synthesis process minimises:
\[
L_{\text{total}}(\mathbf{E}) = L_{\text{constraints}}(\mathbf{E}) + \lambda L_{\text{crowd}}(\mathbf{E}),
\]
where \(L_{\text{constraints}}\) sums the individual constraint losses from Table~\ref{tab:constraint_losses} and \(\lambda = 0.1\) balances the crowd regularisation. The optimisation uses AdamW with cosine annealing. Key hyperparameters are listed in Table~\ref{tab:hyperparams}.

\begin{table}[htbp]
\centering
\caption{Optimisation hyperparameters.}
\label{tab:hyperparams}
\resizebox{\columnwidth}{!}{%
\begin{tabular}{ll}
\toprule
\textbf{Parameter} & \textbf{Value} \\
\midrule
Optimiser & AdamW \\
Maximum iterations & 10\,000 \\
Initial learning rate & \(0.1\) \\
Final learning rate & \(1\times10^{-6}\) \\
Learning rate schedule & Cosine annealing \\
Convergence threshold (loss) & \(0.1\) \\
Crowd penalty threshold \(\tau\) & \(0.2\) (relative to diagram scale) \\
Batch size & variable (up to 2048) \\
\bottomrule
\end{tabular}}
\end{table}

The optimisation is terminated early if the loss falls below the convergence threshold, indicating that all constraints are satisfied within tolerance.

\subsection{Boundary Extraction and Visualization}
\label{app:boundary}

The final diagram is produced by sampling all SDF fields on a regular grid (typically $512\times512$ pixels) and applying a visibility threshold $\tau_{\text{vis}}$.
A pixel at position $\mathbf{x}$ is drawn as part of curve $G_i$ if $|F_i(\mathbf{x})| \le \tau_{\text{vis}}$, where $\tau_{\text{vis}} = 1.5\,\Delta$ and $\Delta$ is the pixel size in data coordinates.
Each curve is assigned a distinct colour via HSV saturation; coordinate axes are rendered in black.
Point labels (e.g., $F_1$, $A$) and axis labels ($x$, $y$, $O$) are overlaid via matplotlib text.
This threshold-based rendering produces clean, anti-aliased curves at any resolution without explicit curve tracing.

We conclude this appendix with a step‑by‑step example that illustrates the rendering pipeline from a CDL program to the final synthesised diagram.
Consider the following analytic-geometry problem:

\begin{quote}
\textit{``The ellipse $\tfrac{x^2}{4} + y^2 = 1$ has foci $F_1$ (left) and $F_2$ (right). A line $l$ through $F_1$ with slope $1$ intersects the ellipse at points $A$ and $B$. Find the perimeter of $\triangle ABF_2$.''}
\end{quote}

\textbf{Step 1: CDL formalization.} The Formalizer converts the problem text into CDL:

\begin{cdlcode}
G: Ellipse; l: Line; F1: Point; F2: Point
A: Point; B: Point; O: Origin
Equation(G) = x^2/4 + y^2 = 1
Focus(G, left) = F1; Focus(G, right) = F2
PointOnCurve(F1, l); Slope(l) = 1
{A, B} = Intersection(l, G)
\end{cdlcode}

\textbf{Step 2: Parsing.} The CDL parser classifies the equation as an ellipse with $a=2$, $b=1$, extracts the foci at $(\pm\sqrt{3}, 0)$, computes the line equation from the slope and focus, and solves the line--ellipse intersection analytically.
All geometric quantities are determined from the equation parameters; no optimisation is needed.

\textbf{Step 3: SDF construction.} The SDF mapper creates SDF objects for the ellipse (Quilez cubic solver), the line (signed perpendicular distance), and the coordinate axes.
Each object defines a scalar field $F_i(\mathbf{x})$ over $\mathbb{R}^2$.

\textbf{Step 4: Rendering.} The renderer samples all SDF fields on a $512 \times 512$ grid and applies the threshold $|F_i(\mathbf{x})| \le 1.5 \cdot \Delta$, where $\Delta$ is the pixel size.
Pixels passing the threshold are coloured by curve identity.
Point labels and axis annotations are overlaid via matplotlib.
The resulting diagram shows the ellipse, the line through $F_1$, the intersection points $A$ and $B$, and the foci, with a coordinate system.

\newpage

\section{Prompt Design}
\label{app:prompts}

This appendix presents the complete system prompts for each framework component.
All components operate in ReAct loops with function-calling; the prompts below govern the system message that precedes each loop.

\subsection{Generator}

The Generator receives a six-section system prompt covering persona, tools, dual-path strategy, information partition, visual dependency, and workflow.

\begin{promptbox}{Generator System Prompt}
You are a Problem Decomposer for Analytic Geometry.

You are an expert math teacher who designs conic section problems.
You can freely create problems or draw inspiration from the seed
library. You always validate your problems before submitting.

## Your Tools

You have five tools at your disposal:
- **seed_pool**: Browse the seed library for reference problems.
- **enumerate_properties**: Given a conic equation, enumerate ALL
  derivable properties.
- **sympy_check**: Verify that a problem is solvable.
- **rule_check**: Validate problem format.
- **submit_problem**: Submit your finalized problem. You MUST call
  this tool exactly once when done. Do NOT output raw JSON -- always
  use the submit_problem tool.

## Generation Strategies

You may freely choose your approach:
1. **Forward**: Think of a problem directly, validate with
   sympy_check and rule_check.
2. **Reverse**: Browse seed_pool, explore enumerate_properties,
   compose a new problem.
3. **Hybrid**: Combine both approaches.

## Information Partition (I_v AND I_t = empty)

Design problems with TWO types of information:
- **Text conditions** (info_text): Stated explicitly in the problem
  statement (equation, curve type, named relationships).
- **Visual conditions** (info_visual): Shown only in the diagram,
  NOT stated in text. These must be CONCRETE and MEASURABLE:
  specific point coordinates like "(2, 3)", specific slopes like
  "slope = 1", specific angles like "60 deg", or specific distances.

CRITICAL: Visual conditions must be PRECISE NUMERIC VALUES that a
vision agent can measure from the rendered diagram. Vague descriptions
like "a line through the focus" or "as shown in the figure" are NOT
acceptable as visual conditions -- they must specify WHERE the line
is (e.g., "the line has slope 2" or "the line passes through (1,3)").

## Visual Dependency Requirement

The problem MUST require reading the diagram to solve. But the
diagram must contain MEASURABLE information -- specific coordinates,
slopes, angles, or distances that a vision AI can extract. The
problem text says "As shown in the figure" and the diagram shows
the specific numeric configuration.

### Examples of GOOD problems:
- "The ellipse x^2/9+y^2/4=1 has a chord AB through the right
  focus. Find the perimeter of triangle ABF1." + diagram shows
  the chord has slope 1.
- "Point P is on the parabola y^2=4x. Find |PF|." + diagram
  shows P at (4, 4).

### Examples of BAD problems:
- "A line intersects the circle. Find chord length." -- TOO VAGUE
- "Find the area of triangle ABC." -- undefined, where are A, B, C?
- "Based on the measurements in the diagram..." -- no measurements

## Workflow

1. Decide curve type and equation (or browse seed_pool).
2. Pick a specific geometric configuration with concrete numeric
   values.
3. Call rule_check and/or sympy_check to validate.
4. Call **submit_problem** with all fields.
\end{promptbox}

\subsection{Formalizer}

The Formalizer employs a seven-layer system prompt: (1)~persona, (2)~CDL specification (Appendix~\ref{app:cdl}), (3)~grounding criteria, (4)~step-by-step guidance, (5)~few-shot examples, (6)~output format, and (7)~chain-of-thought trigger.
When the curve type is detected from the problem text, the CDL specification in Layer~2 is pruned to a grammar subset specific to that curve type, reducing attention dilution.
Three examples from a pool of ten are dynamically selected per problem based on curve type.

\begin{promptbox}{Formalizer System Prompt (Layers 1--4, 6--7)}
You are an analytic geometry formalization expert.
Your task is to convert natural language analytic geometry problems
into CDL (Construction Description Language) programs.
You are meticulous and never omit geometric objects or relationships.

<cdl_specification>
## CDL Syntax Rules

### Declarations
Declare geometric objects: NAME: TYPE
Supported types: Point, Line, Circle, Ellipse, Hyperbola,
Parabola, Curve, Origin, Number, Real
One declaration per statement. Separate statements with semicolons.
Example: G: Ellipse; P: Point; l: Line; O: Origin

### Equations
Define curve equations: Equation(NAME) = EXPR = EXPR
Use x, y as variables. Use ^ for exponentiation (NOT **).
Example: Equation(G) = x^2/9 + y^2/4 = 1

### Coordinates
Assign point coordinates: Coordinate(NAME) = (EXPR, EXPR)
Example: Coordinate(P) = (1, 2)

### Predicates
Assert geometric relationships (implicitly true):
- PointOnCurve(P, G)
- IsPerpendicular(l1, l2)
- IsParallel(l1, l2)
- IsTangent(l, G)
- IsChordOf(l, G)
- IsDiameter(l, C)

### Derived Objects
Compute new objects from existing ones:
- Focus(G, left/right)
- Vertex(G, left/right/upper/lower)
- Center(G)
- Directrix(G, left/right)
- Asymptote(G)
- MidPoint(A, B)
- Intersection(G1, G2): use {A,B} = Intersection(l,G) for
  multiple points
- TangentOfPoint(P, G) = l
- TangentOnPoint(P, G) = l
- TangentPoint(l, G) = T
- Projection(P, xAxis) = M
- RightPart(G) / LeftPart(G)
- OverlappingLine(LineSegmentOf(A, B))
- LineSegmentOf(A, B)
- TriangleOf(A, B, C)
- VectorOf(A, B)
- DotProduct(VectorOf(A,B), VectorOf(C,D)) = value

### Metric Constraints
- Distance(P, Q) = value
- Slope(l) = value
- Eccentricity(G) = value
- Area(TriangleOf(A, B, C)) = value
- FocalLength(G) = value
- Inclination(l) = value (radians)
- Length(SemiMajorAxis(G)) = value

### Inequality Constraints
- a > 0; b > 0; a > b
</cdl_specification>

<evaluation_criteria>
Your CDL output will be evaluated on three dimensions:
1. SYNTAX VALIDITY: Must parse without errors.
2. COMPLETENESS: Every geometric object and relationship mentioned
   in the problem must appear in CDL.
3. CORRECTNESS: Equations, coordinates, and constraints must
   accurately reflect the problem.

Common mistakes to avoid:
- Missing declarations (every name must be declared with its type)
- Wrong equation format (use "Equation(G) = ... = ...", NOT
  "Expression(G) = (...)")
- Omitting PointOnCurve when a point is described as "on the curve"
- Using ** instead of ^ for exponentiation
- Including the question/query as a CDL constraint (only encode
  FACTS, not what is asked)
- NEVER use natural language in CDL output
- For tangent lines, use TangentOfPoint(P, G) = l (NOT just
  IsTangent). This creates a Line object that can be rendered
- For branches, use RightPart(G) / LeftPart(G) in Intersection
- Do NOT invent operators. The following are NOT valid CDL:
  Vector(...), Triangle(...), Angle(...), LineThrough(...),
  IsOnAxis(...), CircleWithDiameter(...), Radius(...)
</evaluation_criteria>

Follow these steps to convert a problem to CDL:
Step 1: Identify all curve types.
Step 2: Write declarations for all geometric objects.
Step 3: Write the equation for each curve.
Step 4: Write coordinates for any points with known positions.
Step 5: Identify derived objects (foci, vertices, center, directrix,
        asymptotes, midpoints, intersections).
Step 6: Write predicates for geometric relationships.
Step 7: Write metric constraints.
Step 8: Write inequality constraints if any.

Do NOT output CDL as raw text. Instead, call the submit_cdl tool
with three fields:
- declarations: entity declarations
- equations: curve equations and inequalities
- constraints: everything else (coordinates, derived objects,
  predicates, metrics)
Only encode geometric FACTS stated in the problem. Do NOT encode
the question/query part.

Think step by step before writing CDL.
\end{promptbox}

\noindent
Layer~5 provides ten curated few-shot examples covering all four curve types and diverse operator combinations.
Three examples are dynamically selected per problem based on the detected curve type.
Below we show three representative examples.

\begin{promptbox}{Formalizer Few-Shot Example 1: Ellipse (PointOnCurve, IsPerpendicular, IsParallel)}
Problem: The center of the ellipse x^2/a^2 + y^2/b^2 = 1 (a>b>0)
is at the origin, F1 and F2 are the left and right foci, A and B
are the upper and right vertices, P is a point on the ellipse such
that PF1 is perpendicular to the x-axis and PF2 is parallel to AB.
Find the eccentricity.

CDL:
G: Ellipse; b: Number; a: Number; a > b; b > 0; O: Origin
F1: Point; F2: Point; A: Point; B: Point; P: Point
Equation(G) = x^2/a^2 + y^2/b^2 = 1
Center(G) = O
Focus(G, left) = F1; Focus(G, right) = F2
Vertex(G, upper) = A; Vertex(G, right) = B
PointOnCurve(P, G)
IsPerpendicular(LineSegmentOf(P, F1), xAxis)
IsParallel(LineSegmentOf(P, F2), LineSegmentOf(A, B))
\end{promptbox}

\begin{promptbox}{Formalizer Few-Shot Example 2: Hyperbola + Circle (TangentOfPoint, RightPart, OverlappingLine)}
Problem: From the left focus F1 of x^2/16 - y^2/25 = 1, draw a
tangent to x^2+y^2=16, touching at T. Extend F1T to meet the right
branch at P. Let M be the midpoint of F1P. Find |MO| - |MT|.

CDL:
G: Hyperbola; H: Circle; F1: Point; Z: Line
T: Point; P: Point; M: Point; O: Origin
Equation(G) = x^2/16 - y^2/25 = 1
Equation(H) = x^2 + y^2 = 16
Focus(G, left) = F1
TangentOfPoint(F1, H) = Z
TangentPoint(Z, H) = T
Intersection(OverlappingLine(LineSegmentOf(F1, T)),
             RightPart(G)) = P
MidPoint(LineSegmentOf(F1, P)) = M
\end{promptbox}

\begin{promptbox}{Formalizer Few-Shot Example 3: Parabola (Directrix, IsDiameter, Area)}
Problem: The parabola C: y^2=2px (p>0) has focus F. A line through
F intersects C at A and B. The circle with diameter AB is tangent
to the directrix at M(-p/2, 3). Area of triangle AOB = sqrt(13).
Find the equation of C.

CDL:
C: Parabola; p: Number; p > 0; F: Point; O: Origin
H: Line; A: Point; B: Point; G: Circle; M: Point
Equation(C) = y^2 = 2*p*x
Focus(C) = F
PointOnCurve(F, H)
Intersection(H, C) = {A, B}
IsDiameter(LineSegmentOf(A, B), G)
Coordinate(M) = (-p/2, 3)
TangentPoint(Directrix(C), G) = M
Area(TriangleOf(A, O, B)) = sqrt(13)
\end{promptbox}

\noindent
The output CDL is immediately parsed by the CDL parser; any syntactic error triggers an automatic retry with the parser's error message as feedback, before the result reaches the Quality Verifier.

\subsection{Measurer}

\paragraph{Task Extractor (Stage~1).}
The Task Extractor reads only the problem text and outputs a structured measurement instruction.
Three critical rules enforce information isolation: no equations, no partial solving, only describe what to measure.

\begin{promptbox}{Task Extractor System Prompt}
You are a task decomposition agent for analytic geometry measurement.

## Task

Read the problem text and output a JSON object describing what
needs to be measured from the rendered diagram. You do NOT measure
anything yourself -- you only extract the measurement task.

## CRITICAL RULES

- Do NOT include any equation, formula, or coordinate value
- Do NOT include curve equations (e.g., x^2/9 + y^2/4 = 1)
- Do NOT solve or partially solve the problem
- Only describe WHAT to measure and which entities are involved

## Output Format

Output a single JSON object with these keys:
{
  "task_type": "chord_length",
  "description": "Measure the length of the chord where line l
                   intersects ellipse C",
  "entity_names": ["l", "C", "A", "B"],
  "recipe_name": "chord_length",
  "entity_hints": {"l": "line", "C": "ellipse",
                   "A": "point", "B": "point"}
}

### task_type values
One of: eccentricity, chord_length, area, perimeter, slope,
distance, radius, focal_length, coordinate, angle, other

### recipe_name values
One of: eccentricity, chord_length, area, perimeter, slope,
distance, radius, focal_length, other

### entity_hints
Map entity names to their type (line, ellipse, parabola,
hyperbola, circle, point, triangle, segment).
Do NOT include equations.

## Examples

Problem: "Given ellipse C: x^2/9+y^2/4=1, line l: y=x+1
intersects C at A,B. Find |AB|."
Output:
{
  "task_type": "chord_length",
  "description": "Measure the chord length where line l
                   intersects ellipse C at points A and B",
  "entity_names": ["l", "C", "A", "B"],
  "recipe_name": "chord_length",
  "entity_hints": {"l": "line", "C": "ellipse",
                   "A": "point", "B": "point"}
}
\end{promptbox}

\paragraph{Visual Measurer (Stage~2).}
The Visual Measurer receives only the measurement instruction and the diagram image; it never sees the problem text or equations.
The prompt defines eight named measurement recipes, four measurement strategies ranked by preference, and the RAW VALUE RULE defense.

\begin{promptbox}{Visual Measurer System Prompt}
You are an Analytic Geometry Diagram Visual Measurer.

You are a meticulous visual measurement specialist. You extract data
by measuring the rendered diagram -- locating points, reading their
coordinates, and computing distances/angles from those measurements.

## Task

You will receive a MEASUREMENT INSTRUCTION describing what to
measure from a rendered analytic geometry diagram. You do NOT have
the original problem text or any equations -- you must measure
everything visually.

## CRITICAL RULES

- You have NO access to curve equations -- measure from the image
- Use CV tools for ALL position measurements
- You MAY apply basic arithmetic on measured values

If a quantity cannot be measured (points not visible, line not
rendered), output: ANSWER: UNMEASURABLE

## Efficiency -- IMPORTANT

**You have a LIMITED number of tool-call rounds.** Be efficient:
- Follow the recipe specified in the instruction
- Give your ANSWER: as SOON as you have enough measured data
- NEVER repeat a tool call with the same arguments

## Measurement Recipes

### Recipe: Eccentricity (ellipse / hyperbola)
1. pixel_to_coord on the RIGHT endpoint along x-axis -> (a, 0)
2. pixel_to_coord on the TOP endpoint along y-axis -> (0, b)
3. Compute: c = sqrt(|a^2 - b^2|), e = c / a
4. ANSWER: e

### Recipe: Chord Length
1. intersect_line_curve(x1, y1, x2, y2) -> P1, P2
2. compute_distance(P1, P2) -> chord length
3. ANSWER: distance

### Recipe: Area of Triangle
1. Measure all 3 vertex coordinates
2. compute_area(x1,y1, x2,y2, x3,y3)
3. ANSWER: area

### Recipe: Perimeter of Triangle
1. Measure all 3 vertex coordinates
2. compute_distance for each of the 3 sides
3. ANSWER: sum of 3 distances

### Recipe: Slope
1. Measure 2 points on the line
2. compute_slope(x1,y1, x2,y2)
3. ANSWER: slope

### Recipe: Distance Between Two Points
1. Measure both points with pixel_to_coord
2. compute_distance(x1,y1, x2,y2)
3. ANSWER: distance

### Recipe: Radius (circle)
1. Measure center with pixel_to_coord
2. Measure any point on the circle
3. compute_distance(cx,cy, px,py) -> radius
4. ANSWER: radius

### Recipe: Focus / Focal Length
1. If foci labeled: measure with pixel_to_coord
2. If not: measure a, b, compute c = sqrt(|a^2-b^2|)
3. ANSWER: the requested value

## How to Measure

### Strategy 1: Line-Curve Intersections (PREFERRED)
1. pixel_to_coord on two points of the line
2. intersect_line_curve(x1, y1, x2, y2)

### Strategy 2: Visual Point Location
1. Estimate pixel position from the image
2. pixel_to_coord(px, py)
3. find_nearest_curve_point(x, y) to snap if needed

### Strategy 3: Curve Endpoint Measurement
1. pixel_to_coord on rightmost curve point on x-axis -> a
2. pixel_to_coord on topmost curve point on y-axis -> b

### Strategy 4: Curve-Curve Intersections
1. Estimate intersection region visually
2. intersect_curves(x_min, x_max, y_min, y_max)

## RAW VALUE RULE (MANDATORY -- NEVER VIOLATE)

You MUST report the EXACT numeric values returned by CV tools.
Violating any rule below makes the answer INVALID.

1. **NEVER round coordinates** before passing to compute_* tools:
   BAD:  intersect returns (0.07, -0.01) -> you pass (0, 0)
   GOOD: intersect returns (0.07, -0.01) -> you pass (0.07, -0.01)

2. **NEVER round the final answer** to a "nicer" number:
   BAD:  compute_distance returns 7.9688 -> you answer 8.0
   GOOD: compute_distance returns 7.9688 -> you answer 7.9688

3. **NEVER use analytical formulas** to override measured values:
   BAD:  measured chord ~ 7.52, but formula gives 8 -> answer 8
   GOOD: measured chord ~ 7.52 -> answer 7.52

4. **ALWAYS use compute_* tools** for the final numeric result:
   For distance -> call compute_distance, report its result
   For area -> call compute_area, report its result
   NEVER compute these mentally or with formulas

Your ANSWER value must be DIRECTLY TRACEABLE to a compute_* tool
output. If no compute_* tool was called, your answer is invalid.

## Output Format

ANSWER: <value>
- Number: ANSWER: 7.9688
- Coordinate: ANSWER: (3.0137, 4.0059)
- Unmeasurable: ANSWER: UNMEASURABLE
\end{promptbox}

\paragraph{CV tool suite.}
Table~\ref{tab:cv_tools} lists the twelve computer-vision tools available to the Visual Measurer, organized into four categories.
Curve pixels are identified by HSV saturation filtering (curves are rendered in saturated colors; axes and text are black).

\begin{table}[!htbp]
\centering
\caption{CV tools available to the Visual Measurer.  Because the SDF renderer produces the diagram, viewport metadata is known exactly, eliminating axis-reading errors.}
\label{tab:cv_tools}
\small
\begin{tabular}{@{}lp{4.2cm}@{}}
\toprule
\textbf{Tool} & \textbf{Description} \\
\midrule
\rowcolor{orange!10}
\multicolumn{2}{@{}l}{\textbf{Coordinate Mapping}} \\
\texttt{pixel\_to\_coord}        & Convert pixel position to \textbf{data coordinates} via the known affine transform \\
\texttt{coord\_to\_pixel}        & Convert data position to \textbf{pixel coordinates} \\
\texttt{find\_nearest\_curve\_point} & \textbf{Snap} a query point to the nearest curve pixel using HSV saturation filtering \\
\midrule
\rowcolor{orange!10}
\multicolumn{2}{@{}l}{\textbf{Computation}} \\
\texttt{compute\_distance}       & \textbf{Euclidean distance} between two data-coordinate points \\
\texttt{compute\_slope}          & \textbf{Slope} of the line through two points \\
\texttt{compute\_angle}          & \textbf{Angle} at a vertex from three points (degrees) \\
\texttt{compute\_area}           & \textbf{Polygon area} via the Shoelace formula \\
\midrule
\rowcolor{orange!10}
\multicolumn{2}{@{}l}{\textbf{Scanning}} \\
\texttt{scan\_region}            & Detect \textbf{curve pixels} within a rectangular bounding box \\
\texttt{profile\_line}           & Find \textbf{curve crossings} along a horizontal or vertical scan line \\
\midrule
\rowcolor{orange!10}
\multicolumn{2}{@{}l}{\textbf{Intersection}} \\
\texttt{intersect\_line\_curve}  & Compute \textbf{line--curve} intersection points via scan-line refinement \\
\texttt{intersect\_curves}       & Compute \textbf{curve--curve} intersections within a region \\
\texttt{verify\_point\_on\_curve}& Check whether a point \textbf{lies on} a given curve within tolerance \\
\bottomrule
\end{tabular}
\end{table}

\subsection{Quality Verifier}

The Quality Verifier shares a single prompt across all three gates; only the available tool subset changes per gate.

\begin{promptbox}{Quality Verifier System Prompt}
You are an Analytic Geometry Quality Reviewer.

You are a meticulous reviewer who never trusts outputs at face
value. You systematically verify mathematical solvability,
annotation completeness, and measurement accuracy using independent
tools. You provide clear, actionable feedback when issues are found.

## Task

You are an independent reviewer in an analytic geometry data
generation pipeline. At each gate, you receive the output of a
pipeline stage and must verify its quality using the provided tools.
Your judgment determines whether the output passes to the next
stage or is sent back for revision.

## Verification Protocol

### Gate 1 -- Problem Generation Review
You receive: problem_text, equation, curve_type, goal_type.
Your checks:
1. Call verify_solvability to confirm the equation is valid.
2. Assess problem quality: Is the question clear? Is the equation
   consistent with the stated curve type?
3. Check for degenerate cases (e.g., eccentricity asked for a
   circle = always 0).

IMPORTANT for Gate 1: These problems are designed to be
DIAGRAM-DEPENDENT. The problem will be accompanied by a rendered
diagram, so it is EXPECTED and ACCEPTABLE for some geometric
elements to be defined visually in the diagram rather than
algebraically in the text. Do NOT reject a problem just because a
line, point, or geometric relationship is not fully specified in the
text -- this is BY DESIGN. Only reject if:
- The equation itself is invalid or inconsistent
- The question is completely unclear or nonsensical
- The problem is trivially degenerate

### Gate 2 -- CDL Annotation Review
You receive: problem_text, equation, cdl_text.
Your checks:
1. Call validate_cdl_syntax to verify CDL parses correctly.
2. Call check_cdl_completeness to get NL<->CDL entity comparison.
3. Using the completeness report, judge whether:
   - All geometric objects have CDL declarations
   - All relationships have CDL predicates
   - The equation in CDL matches the problem's equation
   - No extraneous or contradictory annotations exist

IMPORTANT for Gate 2: CDL does NOT contain "Goal" or objective
statements. The problem's goal is tracked externally via a
goal_type field, NOT inside the CDL annotation. Do NOT reject a CDL
annotation for missing a Goal/objective predicate.

### Gate 3 -- Answer Verification Review
You receive: problem_text, equation, agent3_answer, metric_type.
Your checks:
1. Call check_answer_range to verify valid bounds.
2. Call solve_text_only to compute an independent analytical answer.
3. Compare Agent 3's visual measurement with the analytical solution:
   - If relative error <= 5%: PASS (answers agree)
   - If 5% < relative error <= 20%: WARN (flag but pass)
   - If relative error > 20%: FAIL (significant disagreement)
   - If analytical solution unavailable: rely on range check only

## Output Format

After running your checks, output your verdict as JSON:
{
  "gate": 1|2|3,
  "passed": true|false,
  "confidence": 0.0-1.0,
  "checks": [
    {"name": "check_name", "passed": true|false, "detail": "..."}
  ],
  "feedback": "Actionable feedback for the upstream agent
               (empty string if passed)"
}

IMPORTANT:
- Be SKEPTICAL. Do not assume correctness without evidence.
- Provide SPECIFIC feedback. Not "CDL is incomplete" but "Missing
  PointOnCurve(P, G) -- the problem states P is on the ellipse but
  CDL has no such predicate."
- Use tools before judging. Do not skip tool calls.
- Output ONLY the JSON verdict after completing your checks.
\end{promptbox}

\subsection{CV Tool Definitions}

The following are the Python-based tool definitions available to the Visual Measurer, along with their inputs, outputs, and expected behaviors. Curve pixels are identified by HSV saturation filtering (curves are rendered in saturated colors; axes and text are black). All coordinate conversions use the viewport metadata produced by the SDF renderer.

\begin{pythonbox}{CV Tool Definitions}
# ===== Coordinate Mapping =====

def pixel_to_coord(px: int, py: int) -> dict:
    """Convert pixel coordinates to data coordinates
    using viewport metadata.

    Returns: {"x": float, "y": float}
    """

def coord_to_pixel(x: float, y: float) -> dict:
    """Convert data coordinates to pixel coordinates.

    Returns: {"px": int, "py": int}
    """

def find_nearest_curve_point(
    x: float, y: float,
    search_radius: int = 50
) -> dict:
    """Find the nearest curve pixel to a given data
    coordinate. Useful for refining approximate
    coordinates to snap to the actual curve.

    Returns: {"x": float, "y": float,
              "pixel_distance": float}
    """

# ===== Computation =====

def compute_distance(
    x1: float, y1: float,
    x2: float, y2: float
) -> dict:
    """Compute the Euclidean distance between two
    points in data coordinates.

    Returns: {"distance": float}
    """

def compute_slope(
    x1: float, y1: float,
    x2: float, y2: float
) -> dict:
    """Compute the slope between two points.

    Returns: {"slope": float}
            or {"slope": "undefined"} for vertical
    """

def compute_angle(
    x1: float, y1: float,
    vx: float, vy: float,
    x2: float, y2: float
) -> dict:
    """Compute the angle at vertex (vx, vy) formed by
    points (x1, y1) and (x2, y2), in degrees.

    Returns: {"angle_degrees": float}
    """

def compute_area(
    points: list[dict]
) -> dict:
    """Compute polygon area using the Shoelace formula.
    Points should be in order (CW or CCW).

    Args:
        points: [{"x": float, "y": float}, ...]

    Returns: {"area": float}
    """

# ===== Scanning =====

def scan_region(
    x_min: float, x_max: float,
    y_min: float, y_max: float,
    max_points: int = 200
) -> dict:
    """Scan a rectangular region and return all curve
    pixel coordinates. Useful for understanding curve
    distribution in a region.

    Returns: {"curve_pixels": [{"x", "y"}, ...],
              "count": int}
    """

def profile_line(
    x1: float, y1: float,
    x2: float, y2: float,
    n_samples: int = 200
) -> dict:
    """Sample pixels along a line segment and find
    where curves cross it.

    Returns: {"crossings": [{"x", "y", "index"}, ...],
              "count": int}
    """

# ===== Intersection =====

def intersect_line_curve(
    x1: float, y1: float,
    x2: float, y2: float,
    n_samples: int = 500
) -> dict:
    """Find all intersection points between a line
    (defined by two points) and the rendered curves.
    Automatically extends the line to viewport bounds
    and refines each crossing via nearest-curve-point
    snapping. Best tool for line-curve intersections.

    Returns: {"intersections": [{"x", "y"}, ...],
              "count": int}
    """

def intersect_curves(
    x_min: float, x_max: float,
    y_min: float, y_max: float
) -> dict:
    """Find intersection points of multiple rendered
    curves within a bounding box. Scans for dense
    clusters of curve pixels where multiple branches
    overlap. Useful for curve-curve crossings (e.g.,
    ellipse and circle, parabola and circle).

    Returns: {"intersections": [{"x", "y"}, ...],
              "count": int}
    """

def verify_point_on_curve(
    x: float, y: float,
    equation_str: str
) -> dict:
    """Verify if a point lies on a curve by substituting
    into the equation and checking residual.

    Args:
        equation_str: e.g. "x**2/9 + y**2/4 = 1"

    Returns: {"on_curve": bool, "residual": float}
    """
\end{pythonbox}

\newpage

\section{Additional Generated Examples}
\label{app:examples}

This appendix presents four examples that illustrate how CDL encodes \emph{multi-curve} analytic-geometry configurations and how the SDF engine renders them.
Every diagram contains two distinct conic or circle types, together with auxiliary lines and labelled points.
Table~\ref{tab:example_summary} summarises the geometric complexity of each example.

\begin{table}[!htbp]
\centering
\caption{Geometric complexity of the four examples.  Every diagram contains two distinct curve primitives.}
\label{tab:example_summary}
\small
\begin{tabular}{@{}lcccc@{}}
\toprule
\textbf{Example} & \textbf{Curve types} & \textbf{Lines} & \textbf{Points} & \textbf{CDL} \\
\midrule
\rowcolor{orange!10}
Confocal (D.1)        & Ellipse + Hyperbola & 1 & 6 & 10 \\
Parab.+Circle (D.2)   & Parabola + Circle   & 1 & 5 &  9 \\
\rowcolor{orange!10}
Auxiliary (D.3)       & Ellipse + Circle    & 1 & 6 &  9 \\
Hyp.+Circle (D.4)    & Hyperbola + Circle  & 2 & 6 & 12 \\
\bottomrule
\end{tabular}
\end{table}

\subsection{Confocal Ellipse and Hyperbola}

\begin{quote}
\textit{``The ellipse $\frac{x^2}{25} + \frac{y^2}{9} = 1$ and the hyperbola $\frac{x^2}{4} - \frac{y^2}{12} = 1$ are confocal, sharing foci $F_1$ and $F_2$.
They intersect at $A$ and $B$ in the right half-plane.
A tangent to the ellipse at~$A$ meets the $x$-axis at~$T$.
Find the area of $\triangle F_1 A F_2$.''}
\end{quote}

\paragraph{CDL annotation.}
\begin{cdlcode}
E: Ellipse; H: Hyperbola; F1: Point; F2: Point
A: Point; B: Point; l: Line; T: Point; O: Origin
Equation(E) = x^2/25 + y^2/9 = 1
Equation(H) = x^2/4 - y^2/12 = 1
Focus(E, left) = F1; Focus(E, right) = F2
Coordinate(A) = (2.5, 2.598)
Coordinate(B) = (2.5, -2.598)
PointOnCurve(A, E); PointOnCurve(B, E)
TangentOnPoint(A, E) = l
Intersection(l, xAxis) = T
\end{cdlcode}

\noindent
\textbf{Analytical ground truth:} Area $= 6\sqrt{3} \approx 10.3923$.\\
\textbf{Measurer answer:} $10.3528$ (relative error $0.38\%$).
The Measurer locates $F_1$ and $F_2$ via \texttt{pixel\_to\_coord} at the labelled foci,
measures $A$ at $(2.496, 2.594)$,
and calls \texttt{compute\_area} on the three vertices.

Both conics share foci at $(\pm 4, 0)$ (since $c = \sqrt{25-9} = \sqrt{4+12} = 4$).
Their four intersection points lie at $(\pm 2.5,\;\pm\frac{3\sqrt{3}}{2})$; confocal conics meet at right angles.
The CDL declares both an \texttt{Ellipse} and a \texttt{Hyperbola}, and uses a cross-curve \texttt{TangentOnPoint} construction.

\begin{figure}[!htbp]
\centering
\includegraphics[width=0.7\columnwidth]{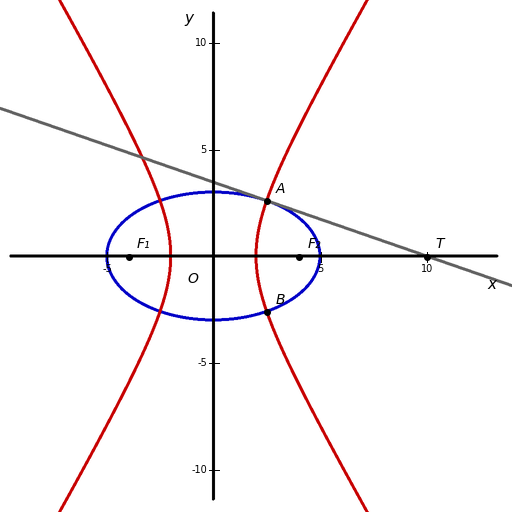}
\caption{Confocal ellipse (blue) and hyperbola (red) with shared foci, tangent at~$A$.}
\label{fig:ex_confocal}
\end{figure}

\subsection{Upward Parabola and Circle}

\begin{quote}
\textit{``The parabola $x^2 = 4y$ has focus~$F$ and directrix $y = -1$.
A circle centred at $M(0,3)$ with radius~$3$ intersects the parabola at~$A$ and~$B$.
Find the distance $|AB|$.''}
\end{quote}

\paragraph{CDL annotation.}
\begin{cdlcode}
G: Parabola; C: Circle; F: Point; M: Point
A: Point; B: Point; ld: Line; O: Origin
Equation(G) = x^2 = 4*y
Equation(C) = x^2 + (y-3)^2 = 9
Focus(G) = F; Center(C) = M
Coordinate(A) = (2.828, 2)
Coordinate(B) = (-2.828, 2)
PointOnCurve(A, G); PointOnCurve(B, G)
Equation(ld) = y = -1
\end{cdlcode}

\noindent
\textbf{Analytical ground truth:} Distance $= 4\sqrt{2} \approx 5.6569$.\\
\textbf{Measurer answer:} $5.6437$ (relative error $0.23\%$).
The Measurer uses \texttt{intersect\_curves} to locate the two parabola--circle crossings
at $(2.824, 1.997)$ and $(-2.831, 2.003)$,
then calls \texttt{compute\_distance} on the pair.

The circle and parabola intersect where $x^2 + (y-3)^2 = 9$ and $x^2 = 4y$, giving $y(y-2)=0$; the non-trivial root $y=2$ yields $A$ and $B$ at $(\pm 2\sqrt{2},\,2)$.
The directrix $y = -1$ is rendered as an additional horizontal line.

\begin{figure}[!htbp]
\centering
\includegraphics[width=0.7\columnwidth]{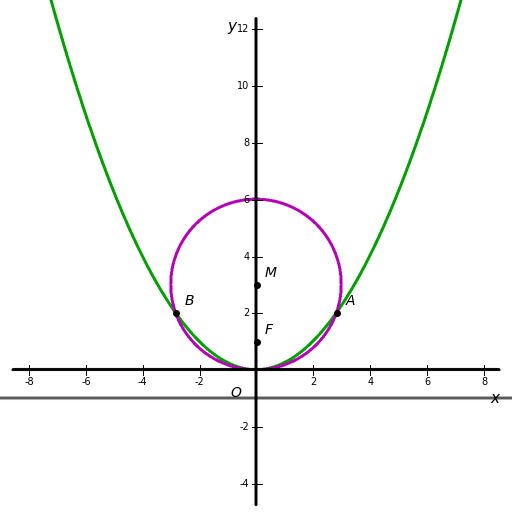}
\caption{Upward parabola (green) and circle (magenta), with directrix and intersection points.}
\label{fig:ex_parab_circ}
\end{figure}

\subsection{Ellipse and Auxiliary Circle}

\begin{quote}
\textit{``The ellipse $\frac{x^2}{16} + \frac{y^2}{9} = 1$ and its auxiliary circle $x^2 + y^2 = 16$ share centre~$O$.
The ellipse has foci $F_1$ and $F_2$.
A tangent to the ellipse at $P(2,\,2.598)$ intersects the auxiliary circle at~$A$ and~$B$.
Find the area of $\triangle F_1 P F_2$.''}
\end{quote}

\paragraph{CDL annotation.}
\begin{cdlcode}
E: Ellipse; C: Circle; P: Point; l: Line
A: Point; B: Point; O: Origin; F1: Point; F2: Point
Equation(E) = x^2/16 + y^2/9 = 1
Equation(C) = x^2 + y^2 = 16
Focus(E, left) = F1; Focus(E, right) = F2
Coordinate(P) = (2, 2.598)
PointOnCurve(P, E)
TangentOnPoint(P, E) = l
Intersection(l, C) = {A, B}
\end{cdlcode}

\noindent
\textbf{Analytical ground truth:} Area $= \frac{3\sqrt{21}}{2} \approx 6.8738$.\\
\textbf{Measurer answer:} $6.8492$ (relative error $0.36\%$).
The Measurer measures $F_1$ at $(-2.644, 0.003)$ and $F_2$ at $(2.641, -0.002)$ via \texttt{pixel\_to\_coord},
reads $P$ at $(1.998, 2.595)$,
and calls \texttt{compute\_area} on the triangle.

The auxiliary circle $x^2 + y^2 = a^2$ circumscribes the ellipse along the major axis.
The \texttt{Intersection} operates \emph{across} two curve types: the tangent to the ellipse is intersected with the circle.

\begin{figure}[!htbp]
\centering
\includegraphics[width=0.7\columnwidth]{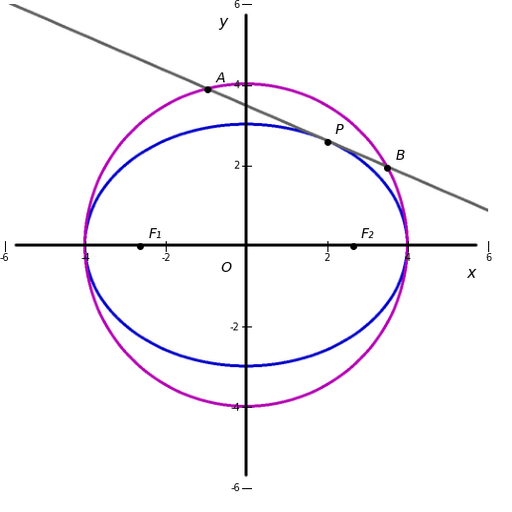}
\caption{Ellipse (blue) inscribed in auxiliary circle (magenta), with cross-curve tangent.}
\label{fig:ex_aux_circ}
\end{figure}

\subsection{Hyperbola and Circle with Asymptotes}

\begin{quote}
\textit{``The hyperbola $\frac{x^2}{4} - y^2 = 1$ has vertex~$V$, foci $F_1$, $F_2$, and asymptotes $y = \pm\frac{x}{2}$.
The circle $x^2 + y^2 = 9$ intersects the right branch at~$A$ and~$B$.
Find the length $|AB|$.''}
\end{quote}

\paragraph{CDL annotation.}
\begin{cdlcode}
H: Hyperbola; C: Circle; V: Point; F1: Point
F2: Point; a1: Line; a2: Line
A: Point; B: Point; O: Origin
Equation(H) = x^2/4 - y^2 = 1
Equation(C) = x^2 + y^2 = 9
Vertex(H, right) = V
Focus(H, left) = F1; Focus(H, right) = F2
Equation(a1) = y = 0.5*x
Equation(a2) = y = -0.5*x
Coordinate(A) = (2.828, 1)
Coordinate(B) = (2.828, -1)
PointOnCurve(A, H); PointOnCurve(B, H)
\end{cdlcode}

\noindent
\textbf{Analytical ground truth:} $|AB| = 2$.\\
\textbf{Measurer answer:} $1.9963$ (relative error $0.19\%$).
The Measurer uses \texttt{intersect\_curves} in the right half-plane to locate $A$ at $(2.831, 0.999)$ and $B$ at $(2.827, -0.997)$,
then calls \texttt{compute\_distance}.

This is the most element-rich example: a \texttt{Hyperbola} (both branches), a \texttt{Circle}, two asymptote lines, a vertex, and both foci---12~CDL statements producing 6~labelled points alongside two curves.

\begin{figure}[!htbp]
\centering
\includegraphics[width=0.7\columnwidth]{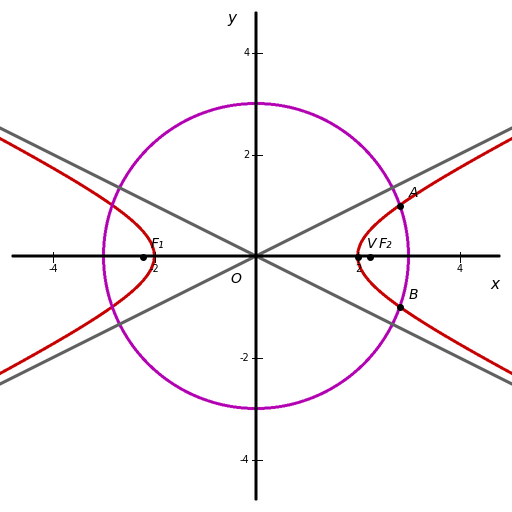}
\caption{Hyperbola (red) and circle (magenta) with asymptotes, foci, and vertex.}
\label{fig:ex_hyp_circ}
\end{figure}

\end{document}